\documentclass[11pt,twoside]{article}

\usepackage[margin=1in]{geometry}
\usepackage{amsmath,amssymb,amsfonts}
\usepackage{graphicx}
\usepackage{xcolor}
\usepackage{hyperref}
\usepackage{natbib} 
\usepackage{blindtext} 

\usepackage{lipsum}
\usepackage{amsfonts}
\usepackage{dsfont}
\usepackage{graphicx}
\usepackage{subfig}
\usepackage{float}
\usepackage{epstopdf}
\usepackage{algorithmicx}
\usepackage{algorithm, algpseudocode}
\usepackage[table,xcdraw]{xcolor}
\usepackage{amsmath, amssymb}
\usepackage{array}
\usepackage{booktabs}
\usepackage{comment}
\usepackage{arydshln}
\usepackage{caption}
\usepackage{multirow}
\usepackage{adjustbox}
\usepackage{threeparttable}

\ifpdf
  \DeclareGraphicsExtensions{.eps,.pdf,.png,.jpg}
\else
  \DeclareGraphicsExtensions{.eps}
\fi

\usepackage{enumitem}
\setlist[enumerate]{leftmargin=.5in}
\setlist[itemize]{leftmargin=.5in}

\definecolor{myblue}{HTML}{1F77B4}
\definecolor{myred}{HTML}{D62728}
\definecolor{mygreen}{HTML}{2CA02C}

\newcommand{\Dset}{\mathcal{D}}

\newcommand{\Xs}{{\color{myred}X}^\mathrm{s}}
\newcommand{\Ys}{{\color{myred}Y}^\mathrm{s}}
\newcommand{\Xv}{{\color{myblue}X}^\mathrm{v}}
\newcommand{\Yv}{{\color{myblue}Y}^\mathrm{v}}
\newcommand{\ms}{{\color{myred}\mu}^\mathrm{s}}
\newcommand{\sigs}{{\color{myred}\sigma}^\mathrm{s}}
\newcommand{\sigv}{{\color{myblue}\sigma}^\mathrm{v}}
\newcommand{\Ps}{{\color{myred}P}^\mathrm{s}}
\newcommand{\Pv}{{\color{myblue}P}^\mathrm{v}}
\newcommand{\ks}{{\color{myred}k}^\mathrm{s}}
\newcommand{\Ks}{{\color{myred}K}^\mathrm{s}}
\newcommand{\Sigs}{{\color{myred}\Sigma}^\mathrm{s}}

\newcommand{\Xr}{{\color{mygreen}X}^\mathrm{r}}
\newcommand{\Yr}{{\color{mygreen}Y}^\mathrm{r}}

\newcommand{\Pre}{{\color{mygreen}P}^\mathrm{r}}

\DeclareMathOperator*{\argmax}{arg\,max} 
\usepackage{amsthm}

\newtheorem{theorem}{Theorem}

\newtheorem{proposition}[theorem]{Proposition}
\newtheorem{lemma}[theorem]{Lemma}
\newtheorem{corollary}[theorem]{Corollary}

\theoremstyle{definition}
\newtheorem{definition}[theorem]{Definition}
\newtheorem{assumption}[theorem]{Assumption}
\newtheorem{example}[theorem]{Example}

\theoremstyle{remark}
\newtheorem{remark}[theorem]{Remark}

\usepackage{amsopn}
\DeclareMathOperator{\diag}{diag}

\title{No-Regret Gaussian Process Optimization\\
of Time-Varying Functions}

\author{
Eliabelle Mauduit\\
Unité de Mathématiques Appliquées\\
ENSTA, Institut Polytechnique de Paris\\
Palaiseau, France\\
\texttt{eliabelle.mauduit@ensta.fr}
\and
Eloïse Berthier\\
U2IS\\
ENSTA, Institut Polytechnique de Paris\\
Palaiseau, France\\
\texttt{eloise.berthier@ensta.fr}
\and
Andrea Simonetto\\
Unité de Mathématiques Appliquées\\
ENSTA, Institut Polytechnique de Paris\\
Palaiseau, France\\
\texttt{andrea.simonetto@ensta.fr}
}

\date{}

\begin{document}

\maketitle
	
	\begin{abstract}%
		Sequential optimization of black-box functions from noisy evaluations has been widely studied, with Gaussian Process bandit algorithms such as GP-UCB guaranteeing no-regret in stationary settings. However, for time-varying objectives, no-regret is unattainable under pure bandit feedback unless strong and often unrealistic assumptions are imposed. We propose a novel method for optimizing time-varying rewards in the frequentist setting, where the objective has bounded RKHS norm almost surely. Time variations are captured through uncertainty injection, enabling heteroscedastic Gaussian process regression that adapts past observations to the current time step. As no-regret is unattainable in general in the strict bandit setting, we relax the latter allowing additional queries on previously observed points. Building on sparse inference and the effect of uncertainty injection on regret, we propose W-SparQ-GP-UCB, an online algorithm that achieves no-regret with a vanishing number of additional queries per iteration. To assess the theoretical limits of this approach, we establish a lower bound on the number of additional queries required for no-regret, proving the efficiency of our method. Finally, we provide a comprehensive analysis linking the temporal regime of the function to achievable regret rates, together with upper and lower bounds on the number of additional queries needed in each regime.
	\end{abstract}
	

	\section{Introduction}
	
	Many real-world systems must be optimized through trial and error, where each trial provides only a noisy glimpse of the true objective. This setting has motivated a large body of work on sequential optimization with noisy feedback, leading to algorithms with strong theoretical guarantees~\citep{Srinivas_2012,Agarwal2011,bubeck2012}. In this setting, a learner seeks to optimize an unknown reward function by iteratively selecting inputs and observing noisy outcomes. Applications span recommender systems~\citep{mary2015}, sensor networks~\citep{roy2016spatio}, and finance~\citep{charpentier2023reinforcement}. A fundamental challenge in this process is balancing \textit{exploration} (sampling points where knowledge about the function is limited), and \textit{exploitation} (selecting inputs already known to yield high rewards).
	
	In the static case, where the objective is fixed over time, bandit algorithms can achieve no-regret guarantees, i.e., asymptotically approach the global optimum~\citep{Srinivas_2012}. In practice, however, many problems are inherently time-varying; for example, traffic routing depends on rush hours, or irrigation depends on weather-driven soil dynamics. In these settings, the goal is no longer to identify a single optimum but to track the optimal trajectory of a function that evolves with time. This makes the problem significantly harder: classical guarantees from the static scenario no longer apply, as past observations become outdated. Several bandit algorithms have recently been proposed to handle non-stationary environments, yet none is able to guarantee accurate asymptotic tracking of the optimal trajectory in terms of regret.
	
	In this work, we propose a new approach to model time variations through a heteroscedastic noise-proxy bound, which quantifies the accuracy of past observations to estimate the current state of the function. We adopt the frequentist Gaussian Process (GP) framework, using GP estimators to approximate the reward function while explicitly accounting for this heteroscedastic noise. Our main contribution is W-SparQ-GP-UCB, an algorithm that leverages a small number of additional queries to refresh outdated information. We show that this mechanism enables accurate tracking of the optimal trajectory with provable no-regret guarantees. Furthermore, we establish both upper and lower bounds on regret across different temporal variations regimes of the objective, as well as upper and lower bounds on the minimal number of additional queries required to recover no-regret.
	
	\subsection{Related work}
	
	Bandit algorithms provide a principled framework for sequential decision-making under uncertainty, where a learner balances exploration and exploitation to maximize cumulative reward. Depending on the nature of the action space, one distinguishes between multi-armed bandits (MAB) with discrete actions~\citep{slivkins2019introduction}, and continuous-action bandits. In the linear case, the expected reward is modeled as a linear function of the action features, and~\cite{dani2008stochastic} established matching upper and lower bounds for the cumulative regret.
	To address nonlinear objectives, linear bandits have been generalized via GP regression, which places a prior distribution over smooth functions through a kernel~\citep{Rasmussen2006Gaussian}. Adaptations of Upper Confidence Bound (UCB) algorithms to this regime naturally trade off exploration and exploitation by maximizing a linear combination of the GP posterior mean and variance. The seminal GP-UCB algorithm of~\cite{Srinivas_2012} achieves no-regret guarantees in both the Bayesian (function sampled from a GP) and frequentist (bounded RKHS norm) settings.
	Subsequent work extended this framework to various real-world contexts, including contextual information~\citep{krause2011contextual}, high-dimensional embeddings~\citep{djolonga2013high}, and temporally evolving objectives~\citep{bogunovic2016,DBLP:Zhoujournals/corr/abs-2102-06296,deng2022weighted}.
	
	In this article, we focus on time-varying reward functions. \cite{bogunovic2016} modeled temporal variations using a GP whose evolution follows a Markov model, quantifying how fast the function changes at every time step. They proposed two approaches to algorithmically capture temporal dynamics: R-GP-UCB, which applies GP-UCB on windows of fixed size, and TV-GP-UCB, which incorporates temporal correlations via a spatio-temporal kernel with a forgetting factor. They established that both algorithms admit linear upper and lower bounds on their regret when the functional change is persistent. For time-varying linear bandits, \cite{cheung2019learning} proposed the SW-UCB algorithm, which achieves sublinear regret by applying GP-UCB on sliding windows of fixed size when the variation budget~$B_T$, defined as the total change of the objective over~$T$ steps, is known and sublinear. \cite{DBLP:Zhoujournals/corr/abs-2102-06296} later derived frequentist regret bounds for both R-GP-UCB and SW-GP-UCB under RKHS smoothness assumptions, recovering results for the linear case. More recently, \cite{deng2022weighted} proposed W-GP-UCB, a weighted-kernel approach that generalizes to both Bayesian and frequentist regimes. Similarly, it guarantees sublinear regret whenever~$B_T$ is known and sublinear.
	
	In these works, time variation is modeled either through variation budgets or forgetting factors, and a common conclusion emerges: when the function evolves persistently (i.e., when~$B_T$ is linear), none of these bandit algorithms provide no-regret guarantees.
	
	\cite{imamura2020timevarying} extended this perspective by allowing non-uniform evaluation times, dynamically adjusting the sampling frequency according to the spatio-temporal kernel's uncertainty structure. Although their method does not achieve sublinear regret for fixed evaluation times and constant forgetting factors, it highlights the deep link between sampling adaptivity and regret control.
	
	A complementary direction was proposed by \cite{Brunzema_2022}, who modeled time variation by injecting uncertainty into the noise variance through a Wiener process. By letting the process variance increase over time, they implicitly capture temporal drift while leveraging standard GP regression. Although they did not provide a regret analysis, their formulation is conceptually close to the weighted-kernel approach, and similar asymptotic bounds are expected to apply.
	
	In the homoscedastic regime, \cite{scarlett2017lower} derived lower bounds for both simple and cumulative regret under the frequentist GP setting, showing that the upper bounds of \cite{Srinivas_2012} are tight for the squared exponential (SE) kernel. \cite{iwazaki2025improved} recently extended these results to heteroscedastic settings by introducing a cumulative variance measure~$V_T$, establishing general lower bounds that hold for arbitrary variance structures. Importantly, their lower bounds become (super)linear for settings where the noise variance follows a prescribed temporal pattern, such as in some of the temporal regimes studied in this paper.
	
	All the above works develop strong regret analyses within the strict bandit framework. Yet, in the time-varying regime, they share a limitation: none guarantees sublinear regret when the function changes at a constant rate (i.e., when~$B_T$ is linear). In~\cite{mauduit2025time}, the bandit feedback is relaxed by allowing the learner to get additional queries at previously observed points at each iteration, yielding~SparQ-GP-UCB, a no-regret algorithm for time-varying Bayesian optimization. This method relies on sparse variational inference in GPs~\citep{burt2020convergence} to select the additional queries locations at each iteration. It is shown that~${\mathcal{O}}(\log^d(T))$ updates per time step, chosen via a Determinantal Point Process (DPP)~\citep{Kulesza_2012}, are sufficient to recover no-regret. Even though this method provides a systematic approach to optimize time-varying objectives, it is still far from the bandit setting as it requires~$\mathcal{O}(T\log^d(T))$ additional queries in total, after~$T$ iterations, where~$d$ is the dimension of the input space. 
	
	\subsection{Contributions}
	In this article, we study time variations in sequential optimization under noisy feedback. We work in the uncertainty-based frequentist setting, where temporal changes in the objective are captured by a single hyperparameter~$\alpha$ (see Assumption~\ref{as.2}), describing how fast the underlying function evolves over time. Depending on~$\alpha$, three fundamentally different regimes naturally emerge:~$\alpha < \frac{1}{3}$ implies slow variations,~$\frac{1}{3} \leq \alpha \leq 1$ indicates moderate variations, while~$\alpha > 1$ represents fast variations. Each regime corresponds to a distinct information–dynamics trade-off, requiring different algorithmic and theoretical tools.
	
	Our main contributions are as follows:
	\begin{enumerate}
		\item \textbf{Algorithmic:} We introduce W-SparQ-GP-UCB, the first algorithm that achieves no-regret in time-varying Gaussian Process bandits while requiring only~$o(1)$ side queries per iteration on average. Assuming~$\alpha$ is known or has a known upper bound, W-SparQ-GP-UCB operates seamlessly across all three regimes.
		\item \textbf{Theoretical guarantees:} We derive regret upper bounds in the strict bandit setting for each regime of temporal variation, generalizing the analysis of \cite{makarova2021}. We then establish improved upper bounds for W-SparQ-GP-UCB, showing that incorporating a vanishing number of side queries suffices to recover sublinear regret. We also provide lower bounds for the regret in the different regimes for the bandit setting, showing the necessity of allowing additional information in general.
		\item \textbf{Fundamental limits:} Using Fano's lemma, we prove lower bounds on the number of side queries necessary to achieve sublinear regret. These results demonstrate the near-optimality of our algorithm in terms of information efficiency, particularly in the fast-variation regime~($\alpha >1$).
	\end{enumerate}
	
{	\begin{table}
		\centering
		\renewcommand{\arraystretch}{1.3}
		\setlength{\tabcolsep}{6pt}
		\caption{Upper and lower regret bounds for time-varying Gaussian Process bandits, where $T$ is the number of rounds the algorithm is run. The regret $R_T$ is formally defined in~\eqref{Eq:CumulativeRegret}.}
		\scalebox{0.8}{\begin{tabular}{|c|c|c|c|}
				\hline
				\textbf{Regime} & \multicolumn{2}{c|}{\textbf{Upper bound on $R_T$}} & \textbf{Lower bound on $R_T$} \\ \cline{2-3}
				& \textbf{Functional form} & \textbf{Comparison to $T$} &  \\ \hline
				$\alpha \geq 1$ 
				& \cellcolor{blue!25} 
				& $\gg T$
				& \cellcolor{blue!25}$\Omega \left( T \right)$ \quad (Prop.~\ref{Prop:BanditLowerRegret}) \\ \cline{1-1} \cline{3-4}
				$\tfrac{1}{3} \le \alpha < 1$ 
				& \cellcolor{blue!25} & $\ge T$ & $\Omega \left( \sqrt{T^{\alpha+1}} \right)$~\citep{iwazaki2025improved} \\ \cline{1-1} \cline{3-4}
				$\alpha < \tfrac{1}{3}$ 
				& \multirow{-3}{*}{\cellcolor{blue!25}$\displaystyle\mathcal{O}\left(\sqrt{d^2T^{3\alpha+1}\log^{2(d+1)}(T)}\right)$\quad (Prop.~\ref{Prop:BanditUpperRegret})} & $o(T)$ 
				& $\displaystyle\Omega\!\big(\sqrt{T}\big)$~\citep{scarlett2017lower} \\ \hline
		\end{tabular}}
		\label{tab:bandit_bounds}
	\end{table}
	\begin{table}
		\centering
		\renewcommand{\arraystretch}{1.3}
		\setlength{\tabcolsep}{6pt}
		\caption{Upper and lower regret bounds as a function of the total number of additional queries~$N_T$ up to a problem parameter dependent constant $C>0$. Parameter~$\tilde{\alpha} < 1/3$ controls the windows' sizes of \textbf{W-SparQ-GP-UCB}.}
		\scalebox{0.63}{
			\begin{tabular}{|c|c|c|c|}
				\hline
				\multirow{2}{*}{\bf Regime} &
				\multicolumn{3}{c|}{\bf $N_T$ (number of additional queries)} \\ \cline{2-4}
				& \cellcolor{blue!25}$N_T = \mathcal{O}\left(T^{\frac{\alpha}{\alpha+1}}\right)$ 
				& \cellcolor{blue!25}$T^{\frac{\alpha}{\alpha+1}} \ll N_T \leq  C T^{\frac{4\alpha-1}{4\alpha}}\log^d T$
				& \cellcolor{blue!25}$N_T \geq C T^{\frac{4\alpha-1}{4\alpha}}\log^d T$ \\ \hline
				
				\multicolumn{4}{|c|}{\textbf{Lower bounds}} \\ \hline
				$\alpha >1$ & \cellcolor{blue!25}$\Omega(T)$ \quad (Thm.~\ref{Thm:LowerBoundSideQueries:formal}) & \cellcolor{blue!25}  & \cellcolor{blue!25} \\  \cline{1-2}
				$\tfrac{1}{3}\le \alpha \le 1$ & \cellcolor{blue!25}$\Omega\left(\sqrt{T}\right)$\quad adapted from~\cite{scarlett2017lower} & \multirow{-2}{*}{\cellcolor{blue!25}$\Omega(\sqrt{T})$\quad adapted from~\cite{scarlett2017lower}}   & \multirow{-2}{*}{\cellcolor{blue!25}$\Omega(\sqrt{T})$\quad adapted from~\cite{scarlett2017lower}}  \\ \hline
				
				\multicolumn{4}{|c|}{\textbf{Upper bound of W-SparQ-GP-UCB}} \\ \hline
				$\alpha > 1$ & \cellcolor{blue!25} Sublinearity unachievable (Thm.~\ref{Thm:LowerBoundSideQueries:formal}) & \multirow{2}{*}{Open question} 
				& \cellcolor{blue!25} \\ \cline{1-2}
				$\tfrac{1}{3}\le \alpha \le 1$ & Open question & & \multirow{-2}{*}{\cellcolor{blue!25}${\mathcal{O}}\left( \sqrt{d^2 T^{3 \tilde{\alpha}+1} \log^{2(d+1)}(T)} \right)$\quad (Thm.~\ref{Thm:regretW-SparQ})}\\ \hline
				
			\end{tabular}
		}
		\label{tab:queries_bounds}
	\end{table}}
	
	We summarize our theoretical results in Table~\ref{tab:bandit_bounds} and Table \ref{tab:queries_bounds}. These tables jointly present our theoretical landscape (we indicate in \colorbox{blue!25}{blue} the principal theoretical contributions of our work). Table \ref{tab:bandit_bounds} shows how regret scales with the number of algorithmic rounds~$T$ under pure bandit feedback, highlighting the transition between slow, moderate, and fast variation regimes. As bandits have been widely studied in the past, some bounds can be directly found in the literature, while some others (the upper bounds in all regimes and the lower bound when~$\alpha >1$) require adaptations to fit our time-varying model. Lower bounds are algorithm-agnostic, whereas the upper-bound analysis is carried out only for GP-UCB under the three variation regimes. Although time-varying GP-UCB variants may yield tighter upper bounds, their regret still scales at least linearly in the general setting, and often has multiplicative polylog terms. Table \ref{tab:queries_bounds} summarizes the need for additional queries. It shows how additional queries reduce regret, and quantifies the minimal number of additional queries~$N_T$ required to maintain no-regret performance when~$\alpha >1$. We finally remark that the division in three time regimes dictated by~$\alpha$ is also novel and key in understanding  time-varying bandits. 
			
	\section{Problem setting}
	
	\subsection{Regularity assumptions on the objective}
	
	We consider the problem of sequentially optimizing the time-varying function~$f : \Dset \times \mathbb{N} \to \mathbb{R}$, where~$\Dset \subset \mathbb{R}^d$ is a compact convex set. At each discrete time step~$t=1,2,\ldots$, a learner selects a point~$x_t \in \Dset$ and observes a noisy evaluation of~$f(x_t,t)$
	\begin{equation}\label{Eq:BaseModel}
		y_t = f(x_t,t) + \epsilon_t, \quad \epsilon_t \sim \mathcal{N}(0, \sigma^2), 
	\end{equation}	
	for some~$\sigma^2 >0$, where the noise sequence~$(\epsilon_t)_{t \geq 0}$ is i.i.d.~Gaussian. For notational convenience, we denote
	\[
	\forall (x,t) \in \Dset \times \mathbb{N}, \quad f(x,t) = f_t(x).
	\]
	The learner's objective is to maximize the sum of rewards~$\sum_{t=1}^T f_t(x_t)$, or equivalently minimize the dynamic cumulative regret,
	\begin{equation}\label{Eq:CumulativeRegret}
		R_T = \sum_{t=1}^T r_t,
	\end{equation}
	where, $r_t$ is the instantaneous regret at step~$t \leq T$:
	\begin{equation}\label{Eq:InstantaneousRegret}
		r_t =  f_t(x_t^*)-f_t(x_t), \qquad x_t^* = \underset{x \in \Dset}{\argmax\,} f_t(x).
	\end{equation}
	
	The overall goal is to derive no-regret algorithms, that is algorithms that achieve a sublinear~$R_T$, for which~$\lim_{T\to\infty} R_T/T = 0$. We focus on online policies, in which the time horizon~$T$ is not fixed a priori. 
	
	\paragraph{Spatial Regularity.} We model each function~$f_t$ as stochastic and we model their spatial smoothness by requiring almost sure (a.s.) membership in the Reproducing Kernel Hilbert Space (RKHS) associated with a continuous bounded kernel~$k$. The RKHS associated with kernel~$k$,~$(\mathcal{H}_k, \langle.,. \rangle_{\mathcal{H}_k})$ is a subset of~$L_2(\Dset)$~\citep{Scholkopf2005KernelMI} such that:
	\begin{equation}\label{Eq:RKHSProp}
		\forall h \in \mathcal{H}_k, \quad \forall x \in \Dset, h(x) = \langle h, k(x,.) \rangle_{\mathcal{H}_k}.
	\end{equation}
	Equation \eqref{Eq:RKHSProp} is called the reproducing property.
	In this article, all results are given for the Squared Exponential (SE) kernel, but can be extended to other standard kernels, such as Matérn. The SE kernel is defined by:
	\begin{equation}\label{Eq:SEKernel}
	k: (x_1, x_2) \in \Dset^2 \mapsto M_k^2 \exp \left(\frac{-\|x_2-x_1\|_2^2}{2l^2}\right),
	\end{equation}
	with~$M_k^2 > 0$ scaling with the data variance and~$l > 0$ the length-scale of the kernel. The SE kernel is said to be bounded in the sense that,
	\begin{equation}\label{Eq:BoundedKernel}
		\forall x \in \Dset, \quad k(x,x) \leq M_k^2.
	\end{equation}
	\begin{assumption}[Spatial regularity]\label{as.1} 
		The sequence~$(f_t)_{t\geq0}$ is a stochastic process taking values in a Reproducing Kernel Hilbert Space with continuous bounded kernel $k$ such that,
			\begin{equation}\label{Eq:RKHS}
				\forall t\mbox{, } \|f_t\|_{\mathcal{H}_k} \leq B \quad \text{almost surely.}
			\end{equation}
	\end{assumption}
	
	Although we do not work under the exact agnostic setting introduced in~\cite{Srinivas_2012}, as we model $f_t$ as a stochastic process, we do not assume a prior on~$f$ as in the Bayesian setting. Hence, the analysis is ultimately frequentist, with guarantees and bounds are conditional to the a.s. event~\eqref{Eq:RKHS}.
	However, because each observation~$(x_t,y_t)$ corresponds to a potentially different function~$f_t$, past data may become obsolete as the function evolves.
	We therefore need a model of temporal evolution that quantifies how older observations can still be trusted at the current step.

    We have chosen not to repeat it at every occasion, but we stress that all the results that are following featuring Assumption~\ref{as.1} are to be thought of to hold at most almost surely. In most cases, the results will be presented to hold with probability~$1-\delta$, which trivially includes an almost sure event. 
	
	\paragraph{Temporal Regularity via Uncertainty Injection.} We introduce an Uncertainty Injection (UI) model to describe temporal drift in~$(f_t)_{t \geq 0}$.
	The idea is to treat past evaluations as noisier, the further back in time they are.
	
	\begin{assumption}[Temporal variability]\label{as.2} 
		There exists a parameter~$\alpha \geq 0$ such that, for all~$t_1 < t_2$ and all~$x \in \Dset$,
		\begin{equation}\label{Eq:TimeVariation}
			f_{t_1}(x) = f_{t_2}(x) + v_{t_1,t_2}(x),
		\end{equation}
		where the noise term~$v_{t_1, t_2}(x)$ is a zero-mean sub-Gaussian random variable with variance~$\sigma^2 (t_2-t_1)^\alpha$, and, for fixed~$t_2$, the~$(v_{t_1,t_2}(x))_{t_1}$  are independent.
	\end{assumption}
	
\begin{figure*}[t!]
    \centering

    \subfloat[$\color{blue}{f_{t_1}(x)+\epsilon_{t_1,t_1}}$]{%
        \includegraphics[width=.48\linewidth]{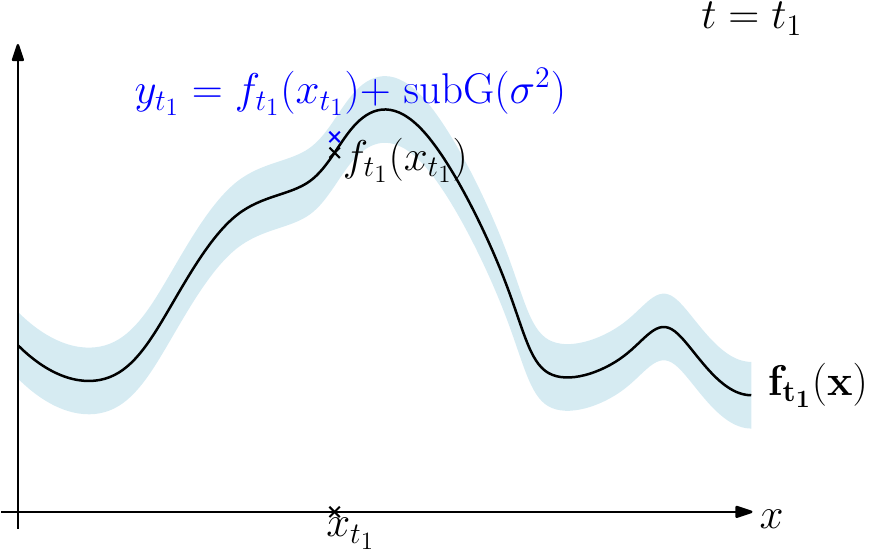}%
        \label{subfig:t1}%
    }\hfill
    \subfloat[$\color{blue}{f_{t_1}(x)+\epsilon_{t_1,t_2}}$]{%
        \includegraphics[width=.48\linewidth]{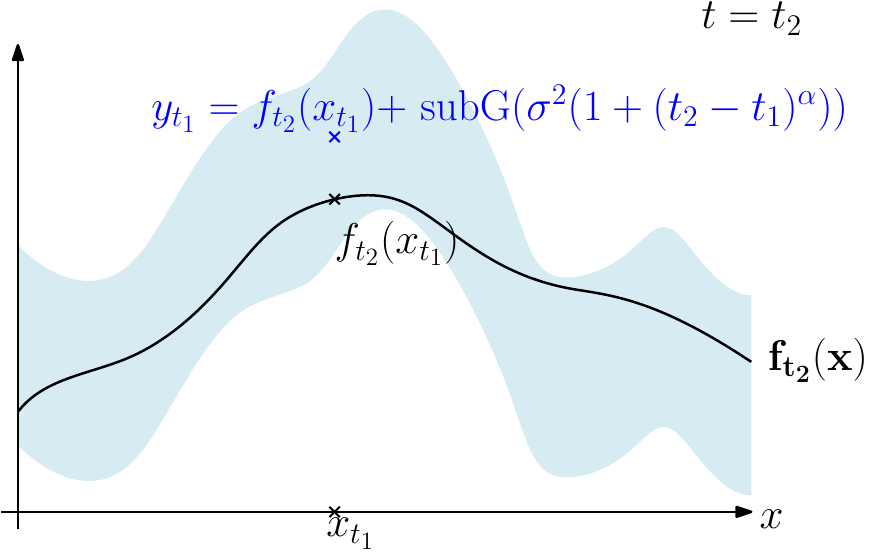}%
        \label{subfig:t2b}%
    }

    \vspace{0.5em}

    \subfloat[$\color{orange}{f_{t_2}(x)+\epsilon_{t_2,t_2}}$]{%
        \includegraphics[width=.48\linewidth]{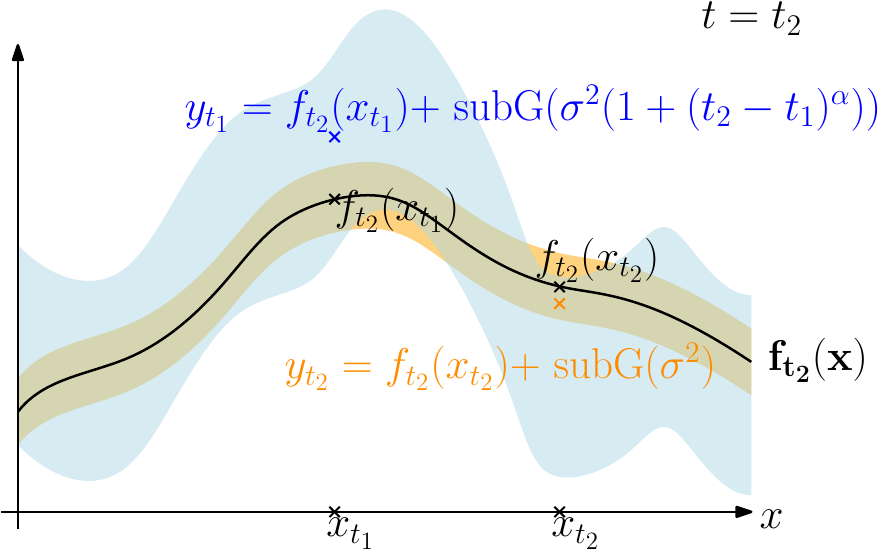}%
        \label{subfig:t2a}%
    }\hfill
    \subfloat[$\color{blue}{f_{t_1}(x)+\epsilon_{t_1,t_3}}, \color{orange}{f_{t_2}(x)+\epsilon_{t_2,t_3}}$]{%
        \includegraphics[width=.48\linewidth]{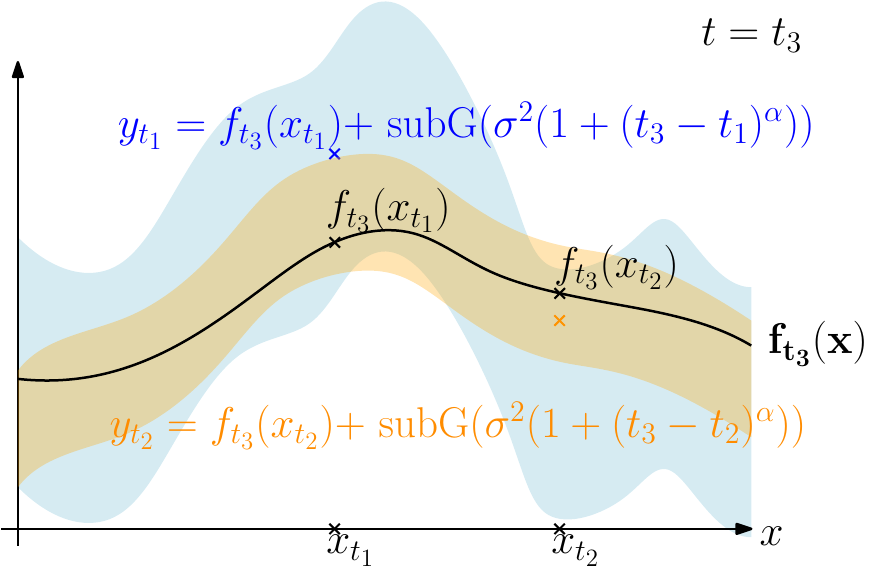}%
        \label{subfig:t3}%
    }

    \caption{Illustration of the uncertainty injection mechanism. The panels display the objective together with observations obtained at times~$t_1$,~$t_2$, and~$t_3$. Each observation is associated with an uncertainty field whose variance increases with the elapsed time according to Assumption~\ref{as.2}. Consequently, uncertainty attached to older observations progressively expands, reflecting the growing gap that arises between past observations and the current objective. The blue uncertainty field, initialized at time~$t_1$, becomes wider at times~$t_2$ and~$t_3$, indicating that the information carried by this observation becomes increasingly uncertain as the objective evolves.}
    \label{fig:UI}
\end{figure*}

	This model captures a wide spectrum of dynamics, from slowly varying drifts~($\alpha$ small) to rapidly changing environments~($\alpha \gg 1$). By leveraging Equation~\eqref{Eq:BaseModel} and Assumption~\ref{as.1}, we obtain the following time-varying model,
	\begin{equation}\label{Eq:TVModel}
		\forall t_1 \leq t_2, \quad y_{t_1} = f_{t_2}(x_{t_1}) + \epsilon_{t_1, t_2},
	\end{equation}
	where~$\epsilon_{t_1,t_2}=v_{t_1, t_2} + \epsilon_{t_1}$ is zero-mean sub-Gaussian with variance~$\sigma^2\left( 1 + (t_2-t_1)^\alpha \right)$, independent across~$t_1$, for fixed~$t_2$. Figure~\ref{fig:UI} provides an intuitive illustration of the uncertainty injection mechanism induced by Assumption~\ref{as.2}. The figure should be interpreted as a conceptual visualization rather than as an exact realization of the stochastic process.

	\begin{example}[Brownian-like drifts ($\alpha=1$)]
    A natural interpretation of Assumption~\ref{as.2} arises when the objective evolves according to a random walk in function space. Assume that, for every~$x \in \Dset$,
    \[
        f_{t+1}(x)=f_t(x)+\xi_t(x),
    \]
    where~$(\xi_t(x))_{t\geq 1}$ are independent centered sub-Gaussian random variables with variance proxy~$\sigma^2$.
    Then, for any~$t_1<t_2$,
    \[
        f_{t_2}(x)-f_{t_1}(x) = \sum_{s=t_1}^{t_2-1}\xi_s(x).
    \]
    Since sums of independent sub-Gaussian random variables remain sub-Gaussian, then the random variable~${f_{t_2}(x)-f_{t_1}(x)}$ is itself sub-Gaussian with variance proxy~$\sigma^2 (t_2-t_1)$, and Assumption~\ref{as.2} holds with~$\alpha=1$.

    This regime corresponds to Brownian scaling: the variance of the temporal drift grows linearly with time. More generally, parameter~$\alpha$ characterizes the rate at which temporal variations accumulate:
    \begin{itemize}
        \item $\alpha<1$: sub-diffusive dynamics, where the objective evolves more smoothly than a random walk;
        \item $\alpha=1$: diffusive (Brownian) dynamics;
        \item $\alpha>1$: super-diffusive dynamics, where temporal variations accumulate faster than a random walk.
    \end{itemize}
\end{example}

	From Assumption~\ref{as.2}, the noise model at~$T$ is~$\rho_T$-sub-Gaussian, with
	\[
	{\rho_T^2 = \sigma^2 \left( 1+(T-1)^\alpha\right)},
	\]
	the variance proxy.
	Combining Assumptions~\ref{as.1} and~\ref{as.2}, we can now perform GP regression in the frequentist setting with a time-weighted covariance structure. Given observations~$(X_T, Y_T) = \{(x_i, y_i)_{i=1}^T\}$, the posterior mean and variance are:
	\begin{align}
		\mu_T(x) &= k_T(x)^\top (K_T + \Sigma_T)^{-1} Y_T, \\
		\sigma_T^2(x) &= k(x,x) - k_T(x)^\top (K_T + \Sigma_T)^{-1} k_T(x),
	\end{align}
	with~$K_T = [k(x_i,x_j)]_{i,j \leq T}$ the observations kernel matrix,~${k_T(x) = [k(x_1,x), \ldots,k(x_T,x)]^\top}$, and the noise matrix~$\Sigma_T = \diag(\text{Var}(\epsilon_{1,T}),\ldots,\text{Var}(\epsilon_{T,T}))$.
	Following the classical GP-UCB strategy~\citep{Srinivas_2012}, the next observation is chosen as:
	\begin{equation}\label{Eq:UCB}
		x_{T+1} = \underset{x \in \Dset}{\arg \max\,} \, \mu_T(x) + \beta_{T+1}\sigma_T(x),
	\end{equation}
	where~$(\beta_t)_{t \geq 0}$ is a sequence of parameters balancing exploration and exploitation. As we will see in the next section, in the regime~${\alpha > 1/3}$, standard UCB strategies may incur growing regret unless additional information is leveraged.

    \paragraph{Relation to the variational budget.}
    In the non-stationary bandit literature (e.g.,~\cite{cheung2019learning,DBLP:Zhoujournals/corr/abs-2102-06296,deng2022weighted}), temporal variations of an RKHS-valued objective are commonly quantified through the variational budget
    \begin{equation}\label{Eq:VarBudget}
        V_T=\sum_{t=1}^{T-1}\|f_{t+1}-f_t\|_{\mathcal{H}_k}.
    \end{equation}
    Under Assumption~\ref{as.1}, $V_T \leq 2(T-1)B$ a.s.  Moreover, 
    \begin{align*} 
    \forall x \in \Dset, \quad \lvert f_{t+1}(x)-f_t(x)\rvert &= \langle f_{t+1}-f_t, k(x,.) \rangle_{\mathcal{H}_k}\\ 
    & \leq \|f_{t+1}-f_t\|_{\mathcal{H}_k} \underbrace{\|k(x,.)\|_{\mathcal{H}_k}}_{\leq M_k},
    \end{align*} 
    and finally
    \[ 
    \|f_T-f_1\|_\infty \leq \sum_{t=1}^{T-1} \|f_{t+1}-f_t\|_\infty \leq V_T\times M_k.
    \]

Hence, Assumption~\ref{as.1} implies that the variational budget is at most linear in~$T$. More importantly, if the temporal drift behaves as a Brownian motion in the RKHS, then  $V_T=\Theta(T)$. Consequently, the regret bounds of~\cite{cheung2019learning, DBLP:Zhoujournals/corr/abs-2102-06296, deng2022weighted} are no longer sublinear.

Assumption~\ref{as.2} provides a more refined description of temporal variability. Rather than controlling the cumulative RKHS increments through a deterministic budget, it models the increments $f_{t_1}-f_{t_2}$ as a centered sub-Gaussian random field whose variance scales as $\sigma^2 |t_2-t_1|^\alpha$. This characterization captures how the magnitude of temporal variations grows relative to the observation-noise scale and allows us to distinguish between different stochastic drift regimes that would all induce a linear variational budget.

	\subsection{Regret bounds in the pure bandit setting}\label{Sec:BanditBounds}
	We are now ready to derive regret bounds under different regimes of~$\alpha$. While some results follow directly from existing literature, our formulation is the first to express them in terms of~$\alpha$, providing a full analysis under our model~\eqref{Eq:TimeVariation}. As we will see in this section, the three regimes emerge from the guarantees provided by regret upper and lower bounds in the bandit setting.
	
	\subsubsection{Upper bounds}
	
	The next proposition derives a general regret upper bound for GP-UCB as a function of the temporal variability parameter~$\alpha$. This result identifies a critical regime transition at~$\alpha = \tfrac{1}{3}$, above which the cumulative regret upper bounds for GP-UCB become superlinear.
	\begin{proposition}[Regret upper bound in the bandit setting]\label{Prop:BanditUpperRegret}
		Let~$k$ be the squared exponential kernel and~$0 < \delta \leq 1$. Consider a sequence of rewards~$(f_t)_{t \geq 0}$ satisfying Assumptions~\ref{as.1} --~\ref{as.2}. Let~$T$ be a time horizon and~$(X_T, Y_T)$ chosen according to GP-UCB (Equation~\eqref{Eq:UCB}), and~$(\beta_t)_{t \geq 0}$ set as 
		\begin{equation}\label{Eq:Beta}   
			\beta_t =  \sqrt{2\log \left( \frac{2\vert \Sigma_t + K_{tt} \vert^{1/2}} {\delta\vert \Sigma_{t}\vert^{1/2}} \right)} + B.  
		\end{equation}  
		Then, with probability at least~$1-\delta/2$, {\em\textbf{GP-UCB}} attains a cumulative regret of 
		\begin{equation}\label{Eq:BanditUpperRegret}
			R_T = \mathcal{O}\left(\sqrt{d^2T^{3\alpha+1}\log^{2(d+1)}(T)}\right).
		\end{equation}
	\end{proposition}
	This result is an extension from Proposition~1 in~\cite{makarova2021}, widening the scope of \textbf{GP-UCB} by showing that the algorithm is still no-regret on objectives with slow variations, i.e., when~${\alpha < 1/3}$ in Equation~\eqref{Eq:TVModel}. The proof relies on the following lemma:
	\begin{lemma}\label{Lem:beta}{(Lemma~7,~\cite{kirschner2018informationdirectedsamplingbandits})}
		Take any~$0 < \delta \leq 1$ and let~$f_T \in \mathcal{H}_k(\Dset)$ and~$\mu_T(.)$ and~$\sigma_T^2(.)$ be the posterior mean and covariance functions of~$f_T(.)$ after observing~$(X_T, Y_T)$ points. Then, for~$(\beta_t)_{t \geq 0}$ defined according to Equation~\eqref{Eq:Beta}, the following holds with probability at least~$1-\delta/2$:
		\begin{equation}\label{Eq:ucb}
			\forall t \in \{1, \hdots, T\}, \forall x \in \Dset, \qquad \lvert \mu_{t-1}(x) - f_t(x) \rvert \leq \beta_t \sigma_{t-1}(x).
		\end{equation}
	\end{lemma}

	\begin{proof}
		The proof follows the same steps as the one of Proposition~1 in~\cite{makarova2021}. We define the upper and lower confidence bounds functions for~$x \in \Dset$:
		\begin{equation*}
			\text{ucb}_t(x) = \mu_{t-1}(x)+\beta_t\sigma_{t-1}(x),\quad 
			\text{lcb}_t(x) = \mu_{t-1}(x)-\beta_t\sigma_{t-1}(x),
		\end{equation*}	
		with~$(\beta_t)_{t \geq 0}$ defined according to Equation~\eqref{Eq:Beta}. 
		We use Lemma~\ref{Lem:beta} (also Lemma~7,~\cite{kirschner2018informationdirectedsamplingbandits}) to bound the instantaneous regret defined as in Equation~\eqref{Eq:InstantaneousRegret} with probability~$1- \delta/2$:
		\begin{equation*}
			r_t \leq \text{ucb}_t(x_t^*)-\text{lcb}_t(x_t) \leq \text{ucb}_t(x_t)-\text{lcb}_t(x_t)= 2 \beta_t\sigma_{t-1}(x_t),
		\end{equation*}
		where the first inequality is obtained by applying Inequality~\eqref{Eq:ucb} twice, once at~$x=x_t$ and once at~$x=x_t^*$. The second inequality derives from the definition of~$x_t$. Then, as the sequence~$(\beta_t)_{t \geq 0}$ is non-decreasing, at horizon~$T$, the cumulative regret is bounded by:
		$$
		R_T \leq 2 \beta_T \sum_{t=1}^T \sigma_{t-1}(x_t).
		$$
		We then follow the exact same steps as \cite{makarova2021} (Appendix A.1.1 Step 4), except that we replace their fixed upper bound~$\bar{\rho}^2$ on the noise variance proxy by our time varying variance-proxy~$\sigma^2(1+(T-1)^\alpha) (\leq \sigma^2 T^\alpha$ for~$T$ large enough):
		\begin{equation}\label{Eq:RegretInter}
			R_T \leq \beta_T \sqrt{2T(1+\sigma^2T^\alpha)\gamma_T},
		\end{equation}
		with~$\gamma_T$ the maximum information gain at~$T$ defined as
		$$
		\gamma_T = \underset{x_1,\ldots,x_T}{\max\,} \log \left( \frac{\Sigma_T + K_{TT}}{\Sigma_T} \right).
		$$
		Following the same computations as \cite{makarova2021} (A.1.2 and A.1.3) with the time-varying proxy~$\sigma^2 T^\alpha$, we obtain the following bounds:
		\begin{equation}\label{Eq:BetaGamma}
			\beta_T = \mathcal{O}\left(\sqrt{\gamma_T}\right) \qquad \text{and} \qquad \gamma_T = \mathcal{O}\left( d T^\alpha \log^{d+1}T \right).
		\end{equation}
		Combining Equations~\eqref{Eq:RegretInter} and \eqref{Eq:BetaGamma}, we obtain with probability $1-\delta/2$ the desired regret result. 
	\end{proof}

	\begin{remark}\label{Rem:weaker-noise-assumption}
		The proof of Proposition~\ref{Prop:BanditUpperRegret} only needs an upper bound on the effective observation variance. Concretely, the fixed variance proxy~$\bar\rho^2$ used in \cite{makarova2021} is replaced here by the time-dependent proxy~$\sigma^2(1+(T-1)^\alpha)\le \sigma^2 T^\alpha$. Therefore, it is sufficient to assume that the noise at time~$t$ is sub-Gaussian with variance proxy~$\rho^2_t$ satisfying~$\rho^2_t = \mathcal{O}(t^{\alpha})$. Under this weaker hypothesis, the bounds on~$\beta_T$ and $\gamma_T$ in the proof remain valid and the regret rate~\eqref{Eq:BanditUpperRegret} is unchanged.
	\end{remark}
	This result highlights a threshold in time-varying bandit optimization, where GP-UCB successfully tracks the optimum when $\alpha < 1/3$. However, for~$\alpha \geq 1/3$, the bound in Proposition~\ref{Prop:BanditUpperRegret} grows faster than~$T$, suggesting that standard bandit feedback might become insufficient to guarantee no-regret. This observation motivates us to relax the strict bandit setting by allowing a limited number of additional queries per round. While Proposition~\ref{Prop:BanditUpperRegret} establishes sufficiency, showing that GP-UCB remains no-regret when~$\alpha < 1/3$, it does not establish necessity. To address this, we turn to a lower bound analysis. In the next section, we provide a characterization of regret lower bounds as a function of~$\alpha$, revealing a regime where additional queries are not merely helpful but necessary. In particular, we show that when~$\alpha \ge 1$, any online algorithm restricted to bandit feedback suffers linear regret, marking a clear boundary between tractable and intractable temporal dynamics.
	
	\subsubsection{Lower bounds}\label{Sec:BanditLowerBounds}
	
	We now complement the upper bound analysis with minimax lower bounds under the pure bandit feedback. These lower bounds quantify the inherent difficulty of tracking a time-varying function and highlight regimes where additional information is necessary. Given a time horizon~$T$, we consider the optimization of a function~$f_T = f(.,T) : \Dset \to \mathbb{R}$, such that  $f_T \in \mathcal{F}_k(B) \text{ a.s.}$, where~$\mathcal{F}_k(B)$ is the RKHS norm ball of the SE kernel~$k$ and of radius~$B$. We assume that the observations follow the time-dependent noise model~\eqref{Eq:TVModel}. For this class of problems, an \emph{algorithm}~$\pi$ is any sequential decision rule that at each time~$t$ selects the next query~$x_t\in\Dset$. The expected cumulative regret of~$\pi$ under target~$f_T$ is~$\mathbb{E}_{f_T}[R_T(\pi)]$, and lower bounds are stated in the minimax sense~($\underset{\pi}{\min\,}\underset{f_T}{\max \,}\mathbb{E}_{f_T}[R_T(\pi)]$). Let us define the cumulative variance at horizon~$T$ as
	\[
	V_T = \sum_{t=1}^T \text{Var}[\epsilon_{t,T}]= \sum_{t=1}^T \sigma^2 (1+(T-t)^\alpha) = \sigma^2 T^{\alpha+1} + o\left(T^{\alpha+1}\right).
	\]
	Intuitively, $V_T$ captures the total uncertainty injected into past observations when estimating the current function~$f_T$. As~$\alpha$ increases, past observations become less informative, making online optimization harder. Using Corollary~6.1 from~\cite{iwazaki2025improved}, which extends the lower regret bound of~\cite{scarlett2017lower} to heteroscedastic noise, the following result holds.
	
	\begin{proposition}[Lower bound in the bandit setting]\label{Prop:BanditLowerRegret}
		For any bandit algorithm, there exists a time-varying GP instance satisfying Assumptions~\ref{as.1} --~\ref{as.2} such that:
		\begin{itemize}
			\item If $0 \leq \alpha < 1$, then, 
		\begin{equation}\label{Eq:BanditLowerRegretSmallAlpha}
			\mathbb{E}[R_T] = \Omega \left( \sqrt{T^{\alpha+1}} \right).
		\end{equation}
			\item If $\alpha \geq 1$: \begin{equation}\label{Eq:BanditLowerRegretLargeAlpha}
			\mathbb{E}[R_T] = \Omega \left( T \right).
		\end{equation}
		\end{itemize}

	\end{proposition}
	
	\begin{proof}
		We consider two cases:
		
		\paragraph{Case~$\alpha < 1$.} In this configuration, the cumulative variance satisfies~$V_T = o(T^2)$, allowing a direct application of Corollary~6.1 in~\cite{iwazaki2025improved} yielding~$\mathbb{E}[R_T] = \Omega \left( \sqrt{T^{\alpha+1} \log^{\frac{d}{2}}T^{1-\alpha}} \right)$, hence~${\mathbb{E}[R_T] = \Omega \left( \sqrt{T^{\alpha+1}} \right)}$. This lower bound generalizes the one from~\cite{scarlett2017lower} to the heteroscedastic setting.
		
		\paragraph{Case~$\alpha \geq 1$.} Here,~$V_T$ scales too rapidly to apply the bound of Corollary~6.1 in~\cite{iwazaki2025improved}, as~$V_T = o(T^2)$ does not hold anymore. We first consider some model, denoted as Model~(1) with Gaussian noise sequence~$\epsilon_t^{(1)}$ with variance~${\sigma_1^2 = \sigma^2(1+(T-t)^{\alpha_1})}$, for~${0<\alpha<1}$. 
		
		Now, let~$\alpha_2 \geq 1$ and~${\sigma_2^2 = \sigma^2(1+(T-t)^{\alpha_2})}$. Then, let~$\eta_t \sim \mathcal{N}(0, \sigma_2^2-\sigma_1^2)$ independent of~$\epsilon_t^{(1)}$ and define,~$\epsilon_t^{(2)} = \epsilon_t^{(1)} + \eta_t$, which is Gaussian with variance~${\sigma_2^2 = \sigma^2(1+(T-t)^{\alpha_2})}$, with~$\alpha_2 \geq 1$. 
		
		We call this new observations model ``Model~(2)''. The coupling realizes Model~(2) as a garbling of Model~(1) and justifies the application of Blackwell’s theorem~\citep{blackwell1953equivalent}. Then, since Model~(1) yields a regret of at least 
		\[
		\mathbb{E}[\sqrt{T^{\alpha_1+1}}] \underset{\alpha_1 \to 1}{\sim} T , 
		\] 
		we conclude for $\alpha_2 \geq 1$, that 
		$
		\mathbb{E}[R_T] = \Omega(T).
		$
		As the class of sub-Gaussian distributions with variance proxy~$\sigma^2(1+(T-t)^{\alpha_2})$ contains the corresponding Gaussians with variance~$\sigma^2(1+(T-t)^{\alpha_2})$, the lower bound established against the Gaussian adversary is also a valid lower bound for the entire sub-Gaussian class.
	\end{proof}
	
	Proposition~\ref{Prop:BanditLowerRegret} fills the last column of Table~\ref{tab:bandit_bounds} and establishes a necessity criterion: for fast-varying objectives~$(\alpha \geq 1)$, standard bandit algorithms cannot achieve sublinear regret. Combined with the sufficiency of GP-UCB when~$\alpha < 1/3$, this motivates the design of W-SparQ-GP-UCB, which leverages a small number of additional queries per iteration to efficiently track the optimum in all regimes. 
	
	\section{The \textbf{W-SparQ-GP-UCB} algorithm}
	
	In this section, we introduce W-SparQ-GP-UCB, an algorithm designed to take over standard GP-UCB methods when~$\alpha \geq 1/3$. This algorithm is guaranteed to be no-regret (see Section~\ref{Sec:SparQUpperBound}) regardless of~$\alpha$ and designed to minimize the number of calls to the expert by leveraging the sufficiency of GP-UCB when~$\alpha < 1/3$. We start by describing the ``expert'' and its role in our method, then we present a naive method called SparQ-GP-UCB which achieves no-regret by greedily calling the expert at each iteration. Finally, we intoduce W-SparQ-GP-UCB, a windowed version of SparQ-GP-UCB and show sub-linear regret guarantees with a vanishing average number of additional queries.
	
	\subsection{Additional queries for sublinear regret}\label{Sec:Expert}
	Motivated by the superlinear regret bounds observed in the pure bandit setting for large values of~$\alpha$ (see Table~\ref{tab:bandit_bounds}), we relax the feedback assumption by assuming access to additional information at each iteration. Specifically, updated evaluations of the function at previously queried points can be obtained at each iteration. However, these additional queries are costly, so they must be carefully selected. To formalize this idea, we introduce the notion of a virtual setting, denoted by the pair~$(\Xv_T=X_T, \Yv_T)$, where for all~$t \leq T$,
	$$
	[\Yv_T]_t = f_T(x_t)+ \epsilon_t, \qquad \epsilon_t \sim \mathcal{N}(0, \sigma^2).
	$$
	In this idealized setting, all observations correspond to the current state of the function~$f_T$, i.e., they are free from temporal drift. Thus, the virtual model coincides with the time-invariant GP regression setting for which GP-UCB is known to achieve sublinear regret \citep{Srinivas_2012}. However, having access to~$(\Xv_T=X_T, \Yv_T)$ at every iteration would imply requesting updates on every past observation at every step, which is clearly impractical. Our goal is to approximate this virtual setting as closely as possible while minimizing the number of additional queries.
	
	\smallskip
	{\bf Approximating the virtual setting sparsely.}
	In~\cite{mauduit2025time}, Corollary~22 from~\cite{burt2020convergence} is extended to the case of heteroscedastic GPs with deterministic inputs~$X_T \subset \Dset$, with~$\Dset$ compact. This result shows that only a small, well-chosen subset of virtual points~$(\Xs_T, \Ys_T)$ is sufficient to approximate~$\Pv_T$, the posterior of the GP regression performed on virtual observations~$(\Xv_T, \Yv_T)$. The sparse inputs~$\Xs_T$ are chosen using an approximate $M$-Determinantal Point Process (DPP)~\citep{kulesza2011k} on the set of past inputs~$X_T$, for some~$M \in \mathbb{N}$. Crucially, the DPP depends only on~$X_T$, not on the virtual outputs~$\Yv_T$ that are unavailable to the learner. It is defined as follows:
	
	\begin{definition}[$M$-Determinantal Point Process]
		Given a positive semidefinite kernel matrix~$K$ of size~${N \geq M}$, an~$M$-DPP is a discrete probability distribution defined over subsets of cardinality~$M$ of the columns of~$K$, by:
		$$\mathrm{Pr}(\mathbf{Z}=Z) = \frac{\mathrm{det}(K_{Z,Z})}{\sum_{\vert Z' \vert = M}\mathrm{det}(K_{Z',Z'})},$$
		where~$K_{Z,Z}$ is the principal submatrix of~$K$ with columns in~$Z \subset \{1, \ldots, N\}^M$.
	\end{definition}
	
	Intuitively, it selects informative and non-redundant locations by giving more weight to submatrices with larger determinants. Since exact sampling from an~$M$-DPP is computationally expensive, we employ the Markov chain sampling algorithm from~\cite{burt2020convergence}, which provides an efficient and accurate approximation.

	\begin{proposition}[Proposition~3.1. in~\cite{mauduit2025time}]\label{Prop:BoundKL}
		Let~$k$ be a SE kernel and ${\Xv_T = X_T}$ the set of inputs chosen by the algorithm. Let~$\Yv_T$ denote the corresponding observations of~$f_T$ with i.i.d. Gaussian noise~$\epsilon_t \sim \mathcal{N}(0, \sigma^2)$. Then, for any~$\eta >0$, there exists an approximation level~${\varepsilon_T = \mathcal{O}\left( \frac{\eta}{T}\right)}$ and a number of additional queries~$Q_T= \mathcal{O}\left( \log^d\left( \frac{T}{\eta} \right)\right)$ such that, if~${(\Xs_T, \Ys_T) \subset (\Xv_T, \Yv_T)}$ is obtained from an~$\varepsilon_T$-approximation of a~$Q_T-DPP$ on $\Xv_T$, the posterior~$\Ps_T$ of the GP regression on~$(\Xs_T, \Ys_T)$ satisfies:
		\begin{equation}\label{Eq:KLThm}
			\mathbb{E}[\mathrm{KL}[\Ps_T\|\Pv_T]] \leq \eta,
		\end{equation}
		where $\mathrm{KL}$ is the Kullback–Leibler divergence.
	\end{proposition}
	
	\smallskip
	Proposition~\ref{Prop:BoundKL} provides a key insight: the virtual, time-invariant GP posterior can be arbitrarily well approximated with only~${\mathcal{O}}(\log^d(t))$ additional queries at each iteration~$t$. Leveraging this mechanism, SparQ-GP-UCB is introduced in~\cite{mauduit2025time}. At each iteration~$t$, it approximates the virtual setting by asking~$Q_t$ additional queries at step~$t$.

	\subsection{SparQ-GP-UCB}
	
	The algorithm operates in rounds, where, at each time step~$t$, a sparse subset of inputs~$\Xs_t$ of cardinal~$Q_t$ is selected via a DPP and updated evaluations~$\Ys_t$ on these inputs are provided. The posterior mean and variance are subsequently computed as:
	\begin{align}
		\ms_t(x) &= (\ks_{t}(x))^\top(\Ks_{t}+\Sigs_t)^{-1} \Ys_t, \label{eq.sbu1}\\
		(\sigs_t)^2(x) &= k(x,x) - (\ks_{t}(x))^\top(\Ks_{t}+\Sigs_t)^{-1} \ks_t(x), \label{eq.sbu2}
	\end{align}
	where~$\Sigs_t = \sigma^2 I_{Q_t}$, and~$\Ks_t \in \mathbb{R}^{Q_t \times Q_t}$,~$\ks_t \in \mathbb{R}^{Q_t}$ denote the kernel matrix and vector evaluated on~$\Xs_t$. The next query point is selected according to the usual UCB rule:
	\begin{equation}\label{ucb.s}
		x_{t+1} = \underset{x \in \Dset}{\argmax}\, \ms_t(x) + \beta_{t+1} \sigs_t(x).
	\end{equation}
	Compared to standard GP-UCB, SparQ-GP-UCB adds two key steps at each iteration:
	\begin{enumerate}
		\item Sparse selection of informative locations via a DPP among the past actions (line~3 in Alg.~\ref{alg:SparQ-GP-UCB}),
		\item Expert feedback that refines the posterior on those sparse points (line~4 in Alg.~\ref{alg:SparQ-GP-UCB}).
	\end{enumerate}
	The resulting procedure is summarized in Algorithm~\ref{alg:SparQ-GP-UCB}:
	
	\begin{algorithm}[H]
		\caption{SparQ-GP-UCB}\label{alg:SparQ-GP-UCB}
		\begin{algorithmic}[1]
			\Require Domain~$\Dset$, kernel~$k$
			\Ensure Sequence of selected actions~$\{x_t\}_{t=1}^T$
			\For{$t=1,2,\hdots$}
			\State Sample~$y_t = f_t(x_t) + \epsilon_{t}$
			\State Perform sparse inference on~$X_t = \{x_1, \ldots, x_t\}$ to obtain locations~$\Xs_t$
			\State Query an expert to obtain updated observations~$\Ys_t$ on~$\Xs_t$ for~$f_t$
			\State Perform Bayesian updates~\eqref{eq.sbu1}-\eqref{eq.sbu2} to obtain~$\ms_t$ and~$\sigs$ using~$(\Xs_t, \Ys_t)$
			\State Choose the next action~$x_{t+1}$ via~\eqref{ucb.s}
			\EndFor
		\end{algorithmic}
	\end{algorithm}
	
	\paragraph{Regret guarantees} In this paragraph, we provide a theoretical regret analysis of SparQ-GP-UCB and show that the latter enjoys sublinear cumulative regret.
	\begin{theorem}[Theorem 1 in \cite{mauduit2025time}: Regret for SparQ-GP-UCB]\label{Thm:regret}
		
		Let~$(f_t)_t$ be a sequence of reward functions and~$y_t = f_t(x_t) + \epsilon_t$, where~$\epsilon_t \sim \mathcal{N}(0, \sigma^2)$ are i.i.d. Under Assumptions~\ref{as.1}--\ref{as.2} and using a SE kernel~$k$, let~$(x_t)_{t=1}^T$ denote the queries of Algorithm~\ref{alg:SparQ-GP-UCB}. Take any~$0 < \delta \leq 1$ and set~$(\beta_t)_{t=1}^T$ as
		\begin{equation}\label{Eq:BetaSparQ}
		\beta_t =  \sqrt{2\log \left( \frac{2\vert \textcolor{red}\Sigma^\mathrm{s}_{t} + \textcolor{red}{K}^\mathrm{s}_{tt} \vert^{1/2}} {\delta\vert \textcolor{red}\Sigma^\mathrm{s}_{t}\vert^{1/2}} \right)} + B .  
		\end{equation} 
		Then, with probability at least $1-\delta$, with~$\mathcal{O}\left(\log^d (t)\right)$ additional queries per round, the cumulative dynamic regret of SparQ-GP-UCB satisfies 
		\begin{equation}\label{Eq:RegretBound}
			R_T = \mathcal{O}\left( \sqrt{\left(\log\left( \frac{1}{\delta} \right) +  d \log^{d+1}\left(d \log \left(T\right) \right)\right)\left( T d \log^{d+1}(T)\right)}\right).
		\end{equation}
	\end{theorem}
	This bound shows that SparQ-GP-UCB achieves sublinear dynamic regret with only a logarithmic number of additional queries per time step. A key aspect is that neither the regret rate nor the total number of additional queries depend on the temporal variation rate~$\alpha$ of the objective. Conversely, it depends on spatial regularity via the DPP. Before giving the proof of Theorem~\ref{Thm:regret}, we recall the following proposition from~\cite{burt2020convergence}: 

	\begin{proposition}\label{Prop:PostError:app}[Proposition~1 of~\cite{burt2020convergence}]
		Consider two posterior distributions~$P$ (exact)  and~$Q$(approximate), with means~$\mu_p$,~$\mu_q$ and variances~$\sigma_p^2$ and~$\sigma_q^2$. Suppose that $2\mathrm{KL}[Q\|P] \leq \eta \leq \frac{1}{5}$ and let~$x \in \mathbb{R}^d$. Then,
		$$\vert \mu_p(x) - \mu_q(x) \vert \leq \sigma_p(x) \sqrt{\eta} \leq \frac{\sigma_q(x)\sqrt{\eta}}{\sqrt{1-\sqrt{3\eta}}} \hspace{10pt} \mbox{and} \hspace{10pt} \lvert 1 - \sigma_q^2(x)/\sigma_p^2(x) \rvert < \sqrt{3\eta}.$$
	\end{proposition}
	
	We now provide key steps and elements for the proof of Theorem~\ref{Thm:regret}. A more detailed version of the proof can be found in~\cite{mauduit2025time}, Sec.~5.2.
	\begin{proof}[Proof of Theorem~\ref{Thm:regret}]
		Let~$R_T$ denote the cumulative regret of SparQ-GP-UCB at horizon~$T$. For a sequence~$(\beta_t)_{t \geq 0}$ defined according to~\eqref{Eq:BetaSparQ}, with probability~$1- \delta/2$:
		\begin{equation}\label{Eq:CumulReg}
			R_T \leq 2 \beta_T\sum_{t=1}^T\sigs_{t-1}(x_t).
		\end{equation}
		If~$\sigv_{t-1}(.)$ is the posterior variance of the GP regression on the virtual set~$(\Xv_{t-1}, \Yv_{t-1})$, combining Proposition~\ref{Prop:PostError:app} with the KL bound (Proposition~\ref{Prop:BoundKL}) yields, with probability~$1-\delta/2$:
		\begin{equation}\label{Eq:CumulSig}
			\sigs_{t-1}(x_t) \leq \sigv_{t-1}(x_t) \sqrt{1+ \sqrt{3\eta}},
		\end{equation}
		taking~$Q_t = \mathcal{O}\left( \log^d \frac{4t}{\delta \eta} \right)$. Combining~\eqref{Eq:CumulReg} and~\eqref{Eq:CumulSig} we obtain,
		\[
		R_T \leq 2 \beta_T \sqrt{1+ \sqrt{3\eta}} \sum_{t=1}^T \sigv_{t-1}(x_t).
		\]
		We recall that the noise variance of observations~$(\Xv_T, \Yv_T)$ is uniform equal to~$\sigma^2$, hence (Proposition~1~\citep{makarova2021}),
		\begin{equation}\label{Eq:CumulReg2}
			R_T = \mathcal{O}(\beta_T \sqrt{T \gamma_T (\sigma^2+1)}).
		\end{equation}	
		Let us now bound~$\beta_T$ and~$\gamma_T$. In SparQ-GP-UCB, the ucb acquisition function is computed using the approximate posterior mean and variance. We thus have: 
		\[
		\beta_T = \sqrt{2\log \left( \frac{2 \vert \textcolor{red}\Sigma^\mathrm{s}_{T} + \textcolor{red}{K}^\mathrm{s}_{TT} \vert^{1/2}} {\delta\vert \textcolor{red}\Sigma^\mathrm{s}_{T}\vert^{1/2}} \right)} + B.
		\]
		By the definition of information gain with the sparse observations (see, e.g.,~\cite{makarova2021}), we have
		$$
		\gamma_{Q_T} \geq \log \left(\frac{\vert \textcolor{red}\Sigma^\mathrm{s}_{T} + \textcolor{red}{K}^\mathrm{s}_{TT} \vert} {\vert \textcolor{red}\Sigma^\mathrm{s}_{T}\vert} \right),
		$$
		so that,
		\begin{equation}\label{Eq:OBeta}
			\beta_T = \mathcal{O}\left(\sqrt{\log\left( \frac{2}{\delta} \right) + \gamma_{Q_T}}\right),
		\end{equation}
		with~$Q_T$ the number of additional queries at step~$T$. Combining~\eqref{Eq:OBeta} and~\eqref{Eq:CumulReg2}, we have a new expression for the regret bound:
		
		\begin{equation}\label{Eq:OReg2}
			R_T = \mathcal{O}\left( \sqrt{\left(\log\left( \frac{2}{\delta} \right) + \gamma_{Q_T}\right)T \gamma_T}\right).
		\end{equation}
		Following the exact same computations as \cite{makarova2021} (Appendix~A.1.3) and using~${Q_T = \mathcal{O}\left( \log^d\left( T\right)\right)}$, we can bound~$\gamma_T$, the information gain about~$f_T$ provided by~$(\Xv_T, \Yv_T)$, and~$\gamma_{Q_T}$, the one provided by~$(\Xs_T, \Ys_T)$, for a SE kernel:
		
		\begin{equation}\label{Eq:Gamma}
			\gamma_T = \mathcal{O} \left( d \log^{d+1}(T) \right),
		\end{equation}
		\begin{equation}\label{Eq:GammaStar}
			\gamma_{Q_T} = \mathcal{O} \left( d \log^{d+1}\left( \log^d \left(T \right) \right) \right)= \mathcal{O} \left( d \log^{d+1} \left(d\log(T) \right) \right).
		\end{equation}
		Finally, if we inject bounds \eqref{Eq:Gamma} and \eqref{Eq:GammaStar} into \eqref{Eq:OReg2}:
		\begin{equation}\label{Eq:OReg3}
			R_T = \mathcal{O}\left( \sqrt{\left(\log\left( \frac{1}{\delta} \right) +  d \log^{d+1}\left(d \log \left(T\right) \right)\right)\left( T d \log^{d+1}(T)\right)}\right).
		\end{equation}
		This proves Theorem~\ref{Thm:regret}.
	\end{proof}
	
	\paragraph{Computational complexity} Now that we have established sublinearity of SparQ-GP-UCB regret, we briefly discuss its computational complexity. 
	
	\begin{corollary}\label{cor:complexity}{(Theorem~2 in~\cite{mauduit2025time})}
		Under the same setting of Theorem~\ref{Thm:regret}, the per-iteration computational complexity of SparQ-GP-UCB is upper bounded by $$\mathcal{O}\left(T^2 \log(T) \log^{3d}\left(\frac{T}{\log(T)}\right)\right).$$
	\end{corollary}	
	Despite requiring expert feedback, SparQ-GP-UCB reduces the cubic cost of full GP-UCB $\mathcal{O}(T^3)$ to nearly quadratic time, since regression is performed on a sparse set at each iteration. This aligns with prior work on scalable GP inference~\citep{leibfried2012tutorial}.

	While SparQ-GP-UCB provides strong theoretical and computational guarantees, its design remains inherently greedy with respect to additional information. At each iteration, the algorithm discards all previously observed outputs and only uses the input locations to run the DPP, relying entirely on the expert to refresh the posterior. As a result, the algorithm maintains sublinear regret only by continuously requesting a logarithmic number of expert queries at every step. Although this guarantees robustness to non-stationarity, it makes SparQ-GP-UCB potentially impractical in real-world applications, where expert feedback may be costly, delayed, or limited in frequency.
	
	To overcome this limitation, we exploit a theoretical insight derived from Proposition~\ref{Prop:BanditUpperRegret}. This result ensures sublinear cumulative regret of GP-UCB whenever the observation noise remains sub-Gaussian with variance proxy~$\sigma^2 T^\alpha$, for~$\alpha<1/3$. The key observation is that this condition does not require the noise to follow a specific time-varying model, it only constrains its growth rate. Hence, if one can ensure that observations remain valid (that is, their noise variance stays within~$\mathcal{O}(T^{\tilde{\alpha}})$, for some~$\tilde{\alpha}<1/3$), these observations can safely be reused across multiple rounds without increasing regret. 
	
	Building on this, we partition time into non-overlapping windows~$(\mathcal{W}_k)_{k \geq 1}$ such that for all~$t \leq \tilde{t} \in \mathcal{W}_k$,
	\[
	\text{Var}(\epsilon_{t, \tilde{t}}) \leq \sigma^2 \tilde{t}^{\tilde{\alpha}}.
	\]
	Within each window, previously collected data remain exploitable, meaning that the noise variance is low enough to guarantee no-regret for GP-UCB. The expert is therefore queried only once per window (at the beginning) after which the algorithm reuses the accumulated data within that window. This approach substantially reduces expert dependence while preserving the no-regret property. The resulting algorithm, called W-SparQ-GP-UCB, is detailed in the next subsection.
	
	\subsection{No-regret with a vanishing mean number of additional queries: W-SparQ-GP-UCB}\label{Sec:SparQUpperBound}
	
\begin{figure}%
    \centering
    \subfloat[\centering Variance proxy without windows]{{\includegraphics[width=.46\linewidth]{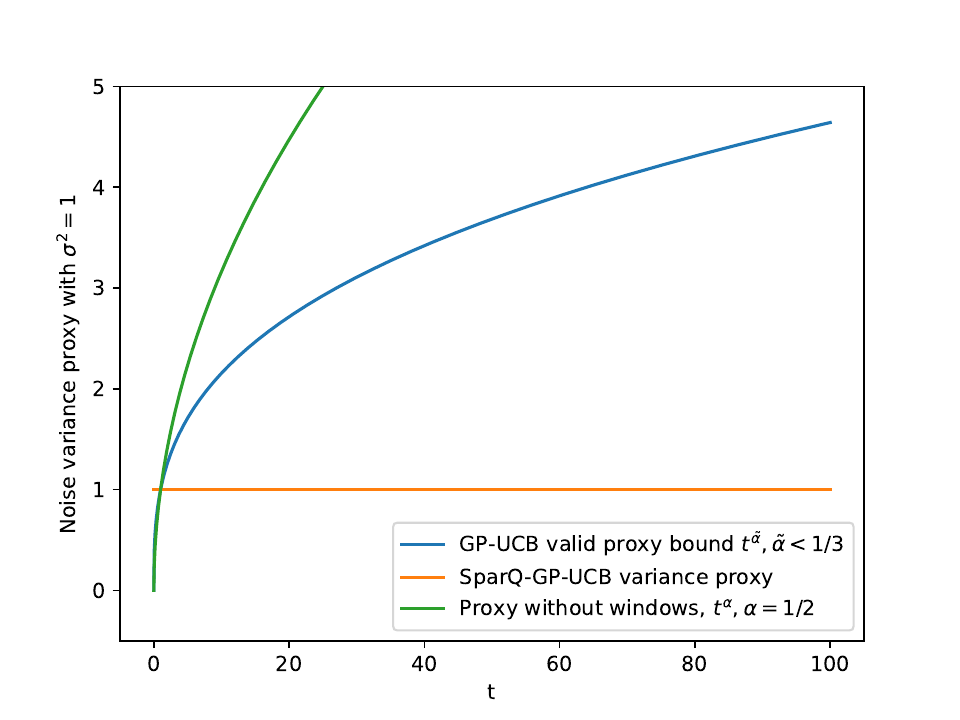} }\label{SubFig:without}}%
    \qquad
    \subfloat[\centering Variance proxy with windows]{{\includegraphics[width=.46\linewidth]{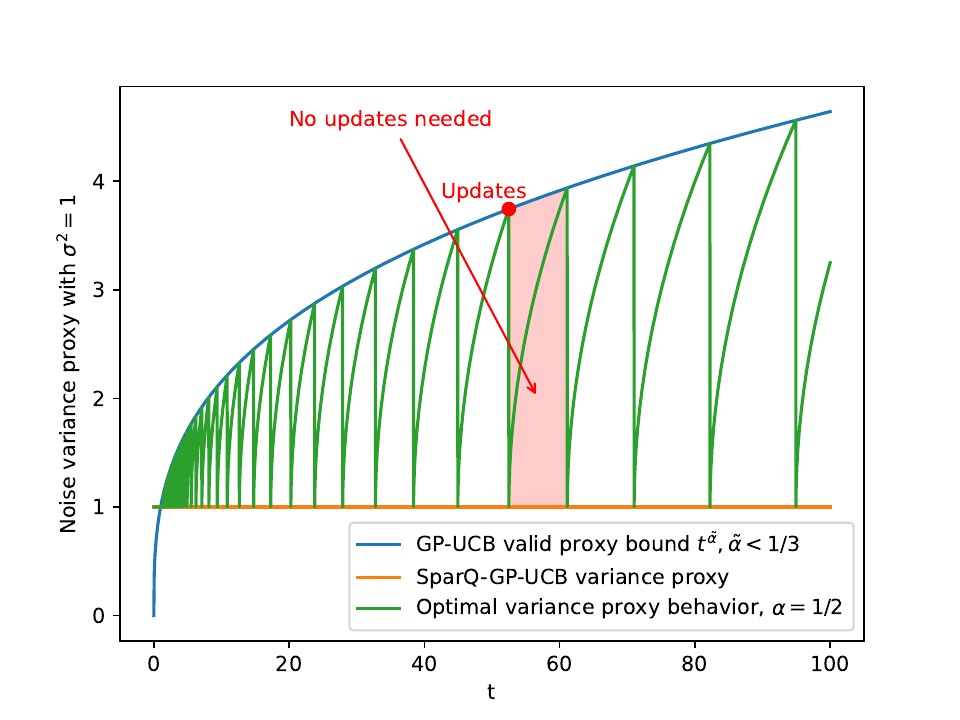} }\label{SubFig:with}}%
    \caption{Illustration of the windowing mechanism. In both panels, the blue curve represents an admissible upper bound on the noise-variance proxy ensuring the no-regret property of GP-UCB. The orange curve corresponds to the variance-proxy bound used by SparQ-GP-UCB, highlighting a significant gap between the admissible threshold and the actual proxy. In Figure~\ref{SubFig:without}, the green curve shows the variance-proxy bound induced by the model~\eqref{Eq:TVModel} with $\alpha=1/2$, which exceeds the admissible threshold. Figure~\ref{SubFig:with} illustrates the proposed windowing strategy: whenever the green curve crosses the blue threshold, additional expert queries are issued to refresh the dataset, resetting the variance proxy to $\sigma^2$. An example of such resulting window is shown as the red shaded region.}%
    \label{Fig:varianceillustration}
\end{figure}

	We now introduce W-SparQ-GP-UCB, a windowed variant of SparQ-GP-UCB that substantially reduces the number of expert queries while preserving sublinear dynamic regret. In SparQ-GP-UCB, the noise-variance proxy is kept constant and equal to~$\sigma^2$. However, Proposition~\ref{Prop:BanditUpperRegret} shows that standard GP-UCB remains no-regret at iteration~$t$ whenever the noise-variance proxy grows as~$o(t^{1/3})$. This observation motivates a windowing strategy: rather than maintaining a single model throughout the entire horizon, we partition time into successive windows such that GP-UCB satisfies the no-regret condition within each window. Whenever the variance-proxy bound exceeds the admissible growth rate, additional expert queries are performed to refresh the dataset and reset the proxy to~$\sigma^2$, thereby starting a new window. The resulting sequence of windows is illustrated in Figure~\ref{Fig:varianceillustration}.

    \subsubsection{Algorithm overview}
    
	Fix~$0 \le \tilde\alpha < 1/3$. We build a sequence of non-overlapping windows~$\mathcal{W}_j = \{t_j,\ldots,t_{j+1}-1\}$ satisfying, for each~$j$,
	\begin{equation}\label{eq:window-size}
		t_j^{\tilde\alpha/\alpha} < t_{j+1}-t_j \le t_j^{\tilde\alpha/\alpha}+1 .
	\end{equation}
	Intuitively, window lengths grow as~$\lfloor t_j^{\tilde\alpha/\alpha}\rfloor$ so that, for any~$t\in\mathcal{W}_j$, noises coming from observations made inside~$\mathcal{W}_j$ have variance at most~$\mathcal{O}(t_j^{\tilde\alpha})$, i.e.,~$\mathcal{O}(t^{\tilde\alpha})$, as~$t_j \leq t$ is the last window start before~$t$.

	At the beginning of each window~$\mathcal{W}_j$ (time~$t_j$) we (i) select a sparse set~$\Xs_{t_j}$ via a~$Q_{t_j}$-DPP (line~3 in Alg.~\ref{alg:W-SparQ-GP-UCB}), (ii) query the expert for updated values~$\Ys_{t_j}$ on~$\Xs_{t_j}$ (line~4 in Alg.~\ref{alg:W-SparQ-GP-UCB}), and (iii) form a posterior using these sparse expert observations (line~5 in Alg.~\ref{alg:W-SparQ-GP-UCB}). For all subsequent rounds~$t\in\mathcal{W}_j$ (lines~8--9 in Alg.~\ref{alg:W-SparQ-GP-UCB}), we run standard GP-UCB using the regression dataset~$(\Xr_t = \Xs_{t_j} \cup \{x_{t_{j+1}},\ldots, x_t\},\Yr_t = \Ys_{t_j} \cup \{y_{t_{j+1}},\ldots, y_t\})$.
	The pseudocode is given in Algorithm~\ref{alg:W-SparQ-GP-UCB}.
	
	\begin{algorithm}
		\caption{W-SparQ-GP-UCB}\label{alg:W-SparQ-GP-UCB}
		\begin{algorithmic}[1]
			\Require Domain \(\Dset\), kernel \(k\), time horizon \(T\), parameters \(\tilde\alpha,\alpha\)
			\Ensure Sequence of selected actions $\{x_t\}_{t=1}^T$
			\For{$t = 1,\dots,T$}
			\If{Beginning of a window}
			\State Perform sparse inference on~$X_t$ to obtain~$\Xs_t$
			\State Get updated outputs~$\Ys_t$
			\State Form posterior by Bayesian update on~$(\Xs_t,\Ys_t)$
			\State Initialize~$(\Xr_t,\Yr_t) \leftarrow (\Xs_t,\Ys_t)$
			\Else
			\State Append~$(x_t,y_t)$ to~$(\Xr_t,\Yr_t)$
			\State Form posterior by Bayesian update on~$(\Xr_t,\Yr_t)$
			\EndIf
			\State Select \(x_{t+1}=\arg\max_{x\in\Dset}\; \mu_t(x)+\beta_{t+1}\sigma_t(x)\)
			\EndFor
		\end{algorithmic}
	\end{algorithm}
    
\begin{figure}[H]
		\centering
		\includegraphics[scale=1]{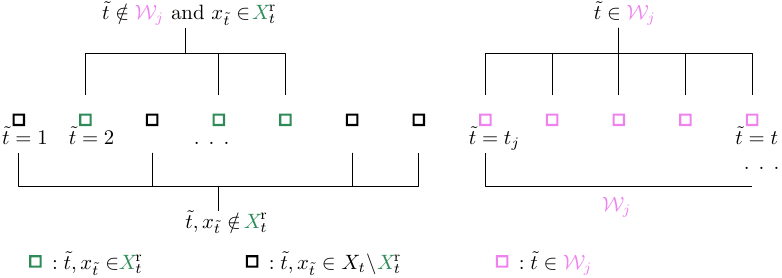}
		\caption{Visualization of the partition of $\{1, \dots, t\}$.}
		\label{Fig:TimesPart}
	\end{figure}

	For the analysis, we define a new virtual dataset~$(\Xv_t,\Yv_t)$ (dependent on the current~$t$ and its window~$\mathcal{W}_j$) so that the noise variance of each virtual observation is bounded by~$2\sigma^2 t^{\tilde\alpha}$. Concretely, for~$t\in\mathcal{W}_j$ and any past index~$\tilde t\le t$, we set the virtual observation variance to
	$$ \mathrm{Var}\big(\epsilon^{\mathrm{v}}_{\tilde t, t}\big) =
	\begin{cases}
		\sigma^2(1+(t-\tilde t)^\alpha), & \text{if } \tilde t\in\mathcal{W}_j,\\[4pt]
		\sigma^2(1+(t-t_j)^\alpha), & \text{if } \tilde t\notin\mathcal{W}_j \text{ and } x_{\tilde t}\in\Xr_t,\\[3pt]
		\sigma^2, & \text{if } \tilde t \notin W_j \text{ and } x_{\tilde t}\notin\Xr_t. 
	\end{cases}
	$$
	For the cases~$\{\tilde t\in\mathcal{W}_j\}$ and~$\{\tilde t\notin\mathcal{W}_j, x_{\tilde t}\in\Xr_t\}$, a query at point~$x_{\tilde{t}}$ has been made at time~$\max(t_j, \tilde{t}) \geq t_j$, by construction of~$\mathcal{W}_j$ and~$\Xr_t$. In that case, the noise variance of the last observation made at input~$x_{\tilde{t}}$ is~$\sigma^2(1+(t-\max(t_j, \tilde{t})^\alpha))\leq 2 \sigma^2 t^{\tilde{\alpha}}$ by Equation~\eqref{eq:window-size}. The remaining possible configuration is~$x_{\tilde{t}} \notin \Xr_t$ (see Figure~\ref{Fig:TimesPart}). In that case, Equation~\eqref{eq:window-size} is not sufficient to conclude that the noise variance of the last observation made at~$x_{\tilde{t}}$ is upper bounded by~$2\sigma^2t^{\tilde{\alpha}}$.This is why we set the noise variance of the associated virtual observations to~$\sigma^2$. 

	The virtual posterior~$\Pv_t$ obtained from~$(\Xv_t,\Yv_t)$ will be used as a reference: we will show that the posterior~$\Pre_t$ built from the algorithm's regressors~$(\Xr_t,\Yr_t)$ is close to~$\Pv_t$ in KL divergence, provided the sparse set is chosen with sufficiently large~$Q_t$.
	
	\subsubsection{Regret and query complexity}
	In this section, we analyze the regret of W-SparQ-GP-UCB, and show that it achieves sublinear regret with a vanishing average number of additional queries per iteration. We also provide an analysis of the algorithm's computational complexity. In order to establish a regret upper bound for W-SparQ-GP-UCB, we start by showing a preliminary result about the Nyström residual trace between the kernel of~$X_T$ and the one of the sparse subset. Below is a definition of the Nyström approximation~\citep{drineas2005nystrom} of a kernel matrix.
	\begin{definition}[Nyström approximation]
		Let~$k$ be a positive definite kernel and~$K_{XX} \in \mathbb{R}^{N \times N}$ be the corresponding kernel matrix over a dataset
		\[
		X = \{x_1, \ldots, x_N\}, \qquad [K_{XX}]_{ij} = k(x_i,x_j).
		\]
		Let~$S = \{x_{s_1}, \ldots, x_{s_M}\} \subset X$ with~$M \leq N$ and define the submatrices:
		$$
		K_{XS} = [k(x_i, x_{s_j})]_{i,j} \in \mathbb{R}^{N \times M}, \qquad K_{SS} = [k(x_{s_i}, x_{s_j})]_{i,j} \in \mathbb{R}^{M \times M}.
		$$
		Then, the Nyström approximation of~$K_{XX}$ based on~$S$ is defined as:
		$$
		U_{XX} = K_{XS} K^\dag_{SS}K_{SX},
		$$
		where $K^\dag_{SS}$ is the pseudoinverse of~$K_{SS}$.
	\end{definition}
	Now, we show the following lemma:
	\begin{lemma}[Monotonicity of Nyström residual trace]\label{Prop:TraceBound}
		Let~$X\subset\Dset$ and~$X^{\mathrm{s}}\subset X$. For any finite~$X^{\mathrm a}\subset\Dset$, denote by~$K$ and~$\tilde K$ the kernel matrices on~$X$ and~$X\cup X^{\mathrm a}$, and by~$U$ and~$\tilde U$ the corresponding Nyström approximations based on queries~$X^{\mathrm s}$ and~$X^{\mathrm s}\cup X^{\mathrm a}$. Then
		\begin{equation}\label{Eq:BoundTrace}
			\mathrm{tr}(\tilde K-\tilde U)\le \mathrm{tr}(K-U).
		\end{equation}
	\end{lemma}
	
	\begin{proof}
		Let~$X = (x_i)_{i=1}^n \subset \Dset$ and~$X^{\mathrm{s}} = (x^s_i)_{i=1}^{n_s} \subset X$. Let~$k : \Dset^2 \to \mathbb{R}$ be the SE kernel with feature map~$\phi : \Dset \to \mathcal{H}_k$ such that,
		\[
		\forall x_1, x_2 \in \Dset \mbox{, } k(x_1,x_2) = \langle \phi(x_1), \phi(x_2)\rangle_{\mathcal{H}_k}.
		\]
		For any sets~$X_1,X_2 \subset \Dset$ we use the notation~$k(X_1,X_2)$ to represent the kernel matrix of sets~$X_1$ and~$X_2$. 
		We start by considering the addition of a singleton~$X^{\mathrm{a}}= \{x_{n+1}\} \in \Dset$, $\tilde{X} = X \cup \{x_{n+1}\}$ and~$\tilde{X}^{\mathrm{s}} = X^{\mathrm{s}} \cup \{x_{n+1}\}$. Let~$K = k(X,X)$ and~$U$ its Nyström approximation based on~$X^{\mathrm{s}}$. We define~$\tilde{K}$ and~$\tilde{U}$ similarly.
		We want to show that~$\mbox{tr}(\tilde{K}-\tilde{U}) \leq \mbox{tr}(K-U)$. Notice that~$K$ is the Gram matrix of vectors~$(\phi(x_i))_{i=1}^n$, i.e,
		\[
		\forall i,j \leq n \mbox{, } K_{i,j} = \langle \phi(x_i), \phi(x_j) \rangle_{\mathcal{H}_k}. 
        \]
        Likewise, 
		\[\forall i,j \leq n+1 \mbox{, } \tilde{K}_{i,j} = \langle \phi(x_i), \phi(x_j) \rangle_{\mathcal{H}_k}.
		\]
		Let~$\mathcal{H}_s = \mbox{Span}\{\phi(x^s_i) \mbox{, } x^s_i \in X^{\mathrm{s}}\}$, $\tilde{\mathcal{H}}^{\mathrm{s}} = \mbox{Span}\{\phi(x^s_i) \mbox{, } x^s_i \in \tilde{X}^{\mathrm{s}}\}$ and~$\mathrm{P}_{\mathcal{H}_{\mathrm{s}}} : \mathcal{H}_k \to \mathcal{H}^{\mathrm{s}}$, $\mathrm{P}_{\tilde{\mathcal{H}}_{\mathrm{s}}} : \mathcal{H}_k \to \tilde{\mathcal{H}}^{\mathrm{s}}$ be the orthogonal projectors on~$\mathcal{H}_s$ and $\tilde{\mathcal{H}}_s$. Note that~$\mathcal{H}^{\mathrm{s}} \subset \tilde{\mathcal{H}}^{\mathrm{s}} \subset \mathcal{H}_k$. Let~$x_i \in X$, then~
		\[
		\exists \alpha_i \in \mathbb{R}^{n_s}, \quad \mathrm{P}_{\mathcal{H}^{\mathrm{s}}}(\phi(x_i)) = \sum_{j=1}^{n_s} \alpha^i_j \phi(x^s_j).
		\]
		The orthogonality conditions can be written as:
		\[
		\forall m \leq n_s \mbox{, } \langle \phi(x_i) - \sum_{j=1}^{n_s} \alpha^i_j \phi(x^s_j), \phi(x^s_m) \rangle_{\mathcal{H}_k} = 0.
		\]
		By using the reproducing property and solving the linear system, we obtain
		\[
		\alpha^i = k(X^{\mathrm{s}}, X^{\mathrm{s}})^{-1}k(X^{\mathrm{s}},x_i),\quad \alpha^i = (\alpha^i_1, \hdots, \alpha^i_{n_s})^\top.
		\]
		From there,
		\begin{align*}
			U_{i,j} &= [k(X,X^{\mathrm{s}})k(X^{\mathrm{s}},X^{\mathrm{s}})^{-1}k(X^{\mathrm{s}},X)]_{i,j} 
			 = (\alpha^i)^\top k(X^{\mathrm{s}}, X^{\mathrm{s}}) \alpha^j \\
			& = \left( \sum_{m=1}^{n_s} \alpha^i_m \langle \phi(x^s_m), \phi(x^s_1) \rangle_{\mathcal{H}_k}, \hdots, \sum_{m=1}^{n_s} \alpha^i_m \langle \phi(x^s_m), \phi(x^s_{n_s}) \rangle_{\mathcal{H}_k}\right) \alpha^j \\
			&= \left( \left \langle \sum_{m=1}^{n_s} \alpha^i_m\phi(x^s_m), \phi(x^s_1)\right \rangle_{\mathcal{H}_k}, \hdots, \left \langle \sum_{m=1}^{n_s} \alpha^i_m \phi(x^s_m), \phi(x^s_{n_s}) \right \rangle_{\mathcal{H}_k}\right) \alpha^j \\
			&= \left( \langle \mathrm{P}_{\mathcal{H}^{\mathrm{s}}}\phi(x_i), \phi(x^s_1) \rangle_{\mathcal{H}_k}, \hdots, \langle \mathrm{P}_{\mathcal{H}^{\mathrm{s}}}\phi(x_i), \phi(x^s_{n_s}) \rangle_{\mathcal{H}_k}\right) \alpha^j \\
			& = \sum_{m=1}^{n_s} \alpha^j_m \langle \mathrm{P}_{\mathcal{H}^{\mathrm{s}}}\phi(x_i), \phi(x^s_m) \rangle_{\mathcal{H}_k} 
			 = \langle \mathrm{P}_{\mathcal{H}^{\mathrm{s}}}\phi(x_i), \mathrm{P}_{\mathcal{H}_s}\phi(x_j) \rangle_{\mathcal{H}_k}.
		\end{align*}
		Similarly, for~$i,j \leq n_s +1 \mbox{, } \tilde{U}_{i,j} = \langle \mathrm{P}_{\tilde{\mathcal{H}}^{\mathrm{s}}}\phi(x_i), \mathrm{P}_{\tilde{\mathcal{H}}^{\mathrm{s}}}\phi(x_j) \rangle_{\mathcal{H}_k}$. Hence, we can write 
		\begin{align*}
			\mbox{tr}(\tilde{K}-\tilde{U}) &= \sum_{i=1}^{n+1} \left( \tilde{K}_{i,i} - \tilde{U}_{i,i} \right) = \sum_{i=1}^{n+1} \left( \|\phi(x_i)\|^2_{\mathcal{H}_k} - \|\mathrm{P}_{\tilde{\mathcal{H}}^{\mathrm{s}}}\phi(x_i)\|^2_{\mathcal{H}_k} \right)\\
			&= \sum_{i=1}^{n+1} \|\phi(x_i) - \mathrm{P}_{\tilde{\mathcal{H}}^{\mathrm{s}}}\phi(x_i)\|^2_{\mathcal{H}_k}.
		\end{align*}
		The last equality is due to the fact that, for~$i \in \{1, \hdots, n+1\}$, $\phi(x_i) = \mathrm{P}_{\tilde{\mathcal{H}}^{\mathrm{s}}}\phi(x_i) +\mathrm{P}_{(\tilde{\mathcal{H}}^{\mathrm{s}})^\perp}\phi(x_i)$, therefore 
		$$\langle \phi(x_i),  \mathrm{P}_{\tilde{\mathcal{H}}^{\mathrm{s}}}\phi(x_i)\rangle_{\mathcal{H}_k} = \|\mathrm{P}_{\tilde{\mathcal{H}}^{\mathrm{s}}}\phi(x_i)\|_{\mathcal{H}_k}^2.$$
		With this in place,
		\begin{align*}
			\mbox{tr}(\tilde{K}-\tilde{U}) &= \sum_{i=1}^{n+1} \|\phi(x_i) - \mathrm{P}_{\tilde{\mathcal{H}}^{\mathrm{s}}}\phi(x_i)\|^2_{\mathcal{H}_k} 
			= \sum_{i=1}^n \|\phi(x_i) - \mathrm{P}_{\tilde{\mathcal{H}}^{\mathrm{s}}}\phi(x_i)\|^2_{\mathcal{H}_k}\\
			& \leq \sum_{i=1}^n \|\phi(x_i) - \mathrm{P}_{\mathcal{H}^{\mathrm{s}}}\phi(x_i)\|^2_{\mathcal{H}_k} = \mbox{tr}(K-U).
		\end{align*}
		The second equality is obtained by noticing that~$\phi(x_{n+1}) \in \tilde{\mathcal{H}}^{\mathrm{s}}$, consequently we obtain
		\[
		{\|\phi(x_{n+1}) - \mathrm{P}_{\tilde{\mathcal{H}}^{\mathrm{s}}}\phi(x_{n+1})\|^2_{\mathcal{H}_k} = 0}.
		\]
		The inequality comes from the fact that~$\mathcal{H}^{\mathrm{s}} \subset \tilde{\mathcal{H}}^{\mathrm{s}}$, so,
		\[
		\forall i \leq n \mbox{, } \|\phi(x_i) - \mathrm{P}_{\tilde{\mathcal{H}}^{\mathrm{s}}}\phi(x_i)\|^2_{\mathcal{H}_k} \leq \|\phi(x_i) - \mathrm{P}_{\mathcal{H}^{\mathrm{s}}}\phi(x_i)\|^2_{\mathcal{H}_k}.
		\]
		We have shown that the trace of the residual matrix is lowered when we add the same point both to the dataset~$X$ and the query set $X^{\mathrm{s}} \subset X$. 
		For general~$X^{\mathrm{a}} = \{x_{n+1}, \ldots, x_{n+k}\}$, we obtain~\eqref{Eq:BoundTrace}  by finite recurrence on~$1 \leq i \leq k$.
	\end{proof}
	We are now ready to show the following result, which establishes a sublinear regret bound for W-SparQ-GP-UCB regardless of~$\alpha$, with a vanishing average number of additional queries~$\frac{N_T}{T}$, where~$N_T = \sum_{t_k \leq T} Q_{t_k}$ is the total number of additional queries. 
	\begin{theorem}[Regret bound for W-SparQ-GP-UCB]\label{Thm:regretW-SparQ}
		Let~$(f_t)_t$ be a sequence of reward functions,~$k$ a squared exponential kernel and~$y_t=f_t(x_t)+\epsilon_t$ with~$\epsilon_t\sim\mathcal{N}(0,\sigma^2)$. Under Assumptions~\ref{as.1}--\ref{as.2}, let~$(x_t)_{t=1}^T$ be the decisions of Algorithm~\ref{alg:W-SparQ-GP-UCB}, for some~$0<\tilde\alpha<1/3$. For any~$0<\delta\le1$, define~$\beta_t$ analogously to~\eqref{Eq:Beta}. Then, with probability at least $1-\delta$, by querying the expert~$Q_{t_k}=\mathcal{O}(\log^d t_k)$ times at each window starting at~$t_k$, the cumulative dynamic regret of W-SparQ-GP-UCB satisfies
		\begin{equation}\label{Eq:WSparQUpper}
			R_T = \mathcal{O}\left(\sqrt{T^{2\tilde{\alpha}+1} d \log^{d+1}T\left(\log \frac{1}{\delta} + d T^{\tilde{\alpha}} \log^{d+1}T\right)}\right).
		\end{equation}
		Moreover, the average number of expert queries per step vanishes: 
		$$
		\frac{N_T}{T} = \mathcal{O}\big(T^{-\tilde\alpha/\alpha}\log^d T\big) = o(1).
		$$
	\end{theorem}
	
	\begin{proof}
		The proof reasoning is similar to the one of Theorem~\ref{Thm:regret}.
		Consider some time horizon~$T$ and~$t_k \leq T$ the last window start before~$T$. Now, let~$(\Xv_{t_k}, \Yv_{t_k})$ be the virtual observations with noise variance proxy~$2\sigma^2 t^{\tilde{\alpha}}$, $(\Xs_{t_k}, \Ys_{t_k}) \subset (\Xv_{t_k}, \Yv_{t_k})$ be the sparse observations obtained from performing a~$Q_{t_k}$-DPP over~$\Xv_{t_k}$, and~$\Pv_{t_k}$,~$\Ps_{t_k}$ the virtual and sparse posteriors. We can bound in probability the KL divergence between~$\Ps_{t_k}$ and~$\Pv_{t_k}$, exactly as it is done in Appendix~B from~\cite{mauduit2025time}, and specifically obtain the following inequality (inequality~(51) from Appendix~B.3 in \cite{mauduit2025time}):
		\[
		\mathbb{E}\big[\mathrm{KL}(\Ps_{t_k}\|\Pv_{t_k})\,\big|\,\Xv_{t_k},\Xs_{t_k}\big] \le \frac{\mathrm{tr}\big(K_{t_kt_k}-U_{t_kt_k}\big)}{\sigma^2},
		\]
		where~$U_{t_kt_k}$ denotes the Nyström approximation of~$K_{t_kt_k}$ using $\Xs_{t_k}$. Both sets~$\Xr_T$ and~$\Xv_T$ are obtained by adding~$\{x_{t_k+1}, \dots, x_T\}$ to~$\Xs_{t_k}$ and~$\Xv_{t_k}$, resp. Applying Lemma~\ref{Prop:TraceBound} yields
		\[
			\mathbb{E}\big[\mathrm{KL}(\Pre_T\|\Pv_T)\,\big|\,\Xr_T,\Xv_T\big] \le \mathbb{E}\big[\mathrm{KL}(\Ps_{t_k}\|\Pv_{t_k})\,\big|\,\Xv_{t_k},\Xs_{t_k}\big] \le \frac{\mathrm{tr}\big(K_{t_kt_k}-U_{t_kt_k}\big)}{\sigma^2} \, .
		\]
		Integrating over the DPP distribution of sparse inputs~$\rho'$:
		\[
		\mathbb{E}\big[\mathrm{KL}(\Pre_T\|\Pv_T)\big] \le \frac{\mathbb{E}_{\rho'}[\mathrm{tr}(K_{t_kt_k}-U_{t_kt_k})]}{\sigma^2} \, .
		\]
		The sparse subset~$\Xs_{t_k}$ is obtained by running an~$\varepsilon_{t_k}$-approximation of a~$Q_{t_k}$-DPP over~$\Xv_{t_k}$. In the proof of Proposition~2 in~\cite{mauduit2025time}, it is shown that, for
		\[
		\varepsilon_{t_k} = \mathcal{O}\left(\frac{\eta}{t_k}\right), \qquad Q_{t_k} = \mathcal{O}\left(\log^d \frac{t_k}{\eta}\right),
		\]
		the following inequality holds:
		\[
		\frac{\mathbb{E}_{\rho'}[\mathrm{tr}(K_{t_kt_k}-U_{t_kt_k})]}{\sigma^2} \leq \eta \, .
		\]
		
		Now, given~$\mathrm{KL}(\Pre_T\|\Pv_T)\le\eta$, Prop.~\ref{Prop:PostError:app} guarantees that 
		\[
		\vert \mu_{\mathrm{v}}(x) - \mu_{\mathrm{r}}(x) \vert \leq \sigma_{\mathrm{v}}(x) \sqrt{\eta} \leq \frac{\sigma_{\mathrm{r}}(x)\sqrt{\eta}}{\sqrt{1-\sqrt{3\eta}}} \hspace{10pt} \mbox{and} \hspace{10pt} \lvert 1 - \sigma_{\mathrm{r}}^2(x)/\sigma_{\mathrm{v}}^2(x) \rvert < \sqrt{3\eta}.
		\]
		An a consequence, the UCBs built from~$\Pre$ are comparable to those from~$\Pv$ up to constants depending on~$\eta$.

		As in the proof of Theorem~\ref{Thm:regret}, we bound the cumulative regret as
		\[ 
		R_T \leq 2 \beta_T \sqrt{1+\sqrt{3\eta}}\sum_{t=1}^T \sigv_{t-1}(x_t).
		\]
		This time, the noise variance of the virtual observations~$(\Xv_T, \Yv_T)$ is bounded by~$2 \sigma^2 T^{\tilde{\alpha}}$ so, by Proposition~1 in  \cite{makarova2021},
		\[
		R_T = \mathcal{O}\left(\beta_T \sqrt{T \gamma_T T^{\tilde{\alpha}}} \right) .
		\]
		Using~$\beta_T=\mathcal{O}(\sqrt{\log(1/\delta)+\gamma_T})$ and the information gain bound $\gamma_T=\mathcal{O}\big(d\,T^{\tilde\alpha}\log^{d+1}T\big)$ for heteroscedastic GP regression with variance proxy~$\mathcal{O}(T^{\tilde{\alpha}})$ and the SE kernel\footnote{Again, obtained by following the exact same steps as in Appendix~A.1.3 from~\cite{makarova2021}.} yields the claimed regret rate
		\[
		R_T = \mathcal{O}\left(\sqrt{T^{2\tilde{\alpha}+1} d \log^{d+1}T\left(\log \frac{1}{\delta} + d T^{\tilde{\alpha}} \log^{d+1}T\right)}\right). 
		\] 
		Let us now bound the average number of additional queries per iteration~$N_T/T$.
		Let~$n(T)$ be the number of windows up to time~$T$, then~$t_{n(T)} \leq T < t_{n(T)+1}$. Rewriting Equation~\eqref{eq:window-size} as
		\[
		\frac{t_{j+1}-t_{j}}{t_{j}^{\tilde{\alpha}/\alpha}} - t_{j}^{-\tilde{\alpha}/\alpha} \leq 1 < \frac{t_{j+1}-t_{j}}{t_{j}^{\tilde{\alpha}/\alpha}},
		\]
		and summing for~$j=1,\dots,n(T)$, we obtain
		$$n(T) = \Theta\left( \sum_{j=1}^{n(T)} \frac{t_{j+1}-t_{j}}{t_{j}^{\tilde{\alpha}/\alpha}} \right).$$
		Since~$t \mapsto t^{-\tilde{\alpha}/\alpha}$ is decreasing on $[1,\infty)$, Equation~\eqref{eq:window-size} gives, for~$j \geq 2$:
		\[
		\int_{t_j}^{t_{j+1}} t^{-\tilde{\alpha}/\alpha}\, \mathrm dt \le t_j^{-\tilde{\alpha}/\alpha}(t_{j+1}-t_j)
		\le \int_{t_{j-1}}^{t_j} t^{-\tilde{\alpha}/\alpha}\, \mathrm dt.
		\]
		Again, summing for $j=2,\dots,n(T)$ gives
		\[
		\sum_{j=2}^{n(T)} \frac{t_{j+1}-t_j}{t_j^{\tilde{\alpha}/\alpha}}
		= \Theta \left( \int_{t_1}^{t_{n(T)+1}} t^{-\tilde{\alpha}/\alpha}\, \mathrm dt \right).
		\]
		The contribution of the missing term~$\frac{t_2-t_1}{t_1^{\tilde\alpha/\alpha}}$ is a fixed constant, hence adding it does not affect the asymptotics of the sum. Therefore
		 \begin{align*}
		 n(T) &= \Theta \left( \int_{t_0}^{t_{n(T)}} t^{-\tilde{\alpha}/\alpha}\,dt \right)\\
		 & = \Theta(T^{(\alpha-\tilde{\alpha})/\alpha}).
		 \end{align*}
		Then,
		\[
		\frac{N_T}{T} \lesssim \frac{n(T)\log^d (n(T))}{T} = \mathcal{O}\big(T^{-\tilde\alpha/\alpha}\log^d T\big).
		\]
		This completes the proof.
	\end{proof}
	
	\subsubsection{Computational complexity}
	The computational complexity of SparQ-GP-UCB (Corollary~\ref{cor:complexity})~was given per iteration. However, here, we reason ``per-window'' because the complexity of the step widely depends on whether or not we are at the start of a window. For~$k \in \mathbb{N}$, the cost of window~$\mathcal{W}_k$ is given by the sum of~\textbf{(i)} the~$Q_{t_k}$-DPP at time~$t_k$ and~\textbf{(ii)} all the posterior updates at steps~$t_k, \ldots, t_{k+1}-1$.
	
	{\bf Complexity of the DPP.} As in SparQ-GP-UCB, for our choice of~${Q_{t_k}=\mathcal{O}(\log^d t_k)}$, the computational complexity of the DPP is
	$$\mathcal{O}\big(t_k Q_{t_k}^3(\log\log t_k + \log Q_{t_k} + \log(1/\varepsilon_T^2))\big).$$
	
	{\bf Complexity of the posterior updates.} For the~$\lfloor t_k^{\tilde{\alpha}/\alpha}\rfloor$ rounds in~$\mathcal{W}_k$, the cost of the posterior updates is dominated by
	\[
	\mathcal{O}\Big( \underbrace{t_k^{\tilde{\alpha}/\alpha}}_{\text{number of rounds}} \underbrace{(Q_{t_k}+t_k^{\tilde{\alpha}/\alpha})^2(t_k + t_k^{\tilde{\alpha}/\alpha})}_{\textbf{highest-cost regression}} ~~ \Big) \, .
	\]

	Hence, the cost of each window is dominated by the cost of the~$\lfloor t_k^{\tilde{\alpha}/\alpha}\rfloor$ regressions, which is~$\mathcal{O}\Big( t_k^{\frac{\tilde{\alpha} + \alpha}{\alpha}} (\log^d(t_k)+t_k^{\tilde{\alpha}/\alpha})^2\Big)$. At step~$T$, there have been~$\mathcal{O}\left(T^{\frac{\alpha - \tilde{\alpha}}{\alpha}}\right)$ windows, giving a total complexity of~$\mathcal{O}\left(T^2(\log^d(T)+T^{\tilde{\alpha}/\alpha})^2\right)$. This is still better than the \emph{total} computational cost of GP-UCB~($\mathcal{O}\left(T^4\right)$), due to the fact that~$\tilde{\alpha}<1/3 \leq \alpha$ (as we only consider~$\alpha \geq 1/3$, since GP-UCB is no regret for~$\alpha < 1/3$).
	\begin{remark}[Choice of parameter~$\tilde{\alpha}$]
		Here, the hyperparameter~$\tilde{\alpha} < 1/3$ controls the windows' sizes. The closer it is to~$1/3$, the fewer additional queries are made. In that sense, larger~$\tilde{\alpha}$ are preferred. However, if the priority is on the computational complexity of the algorithm, smaller~$\tilde{\alpha}$ must be prioritized.
	\end{remark}

	\section{Numerical results}
	
	We evaluate the performance of our method on both synthetic and real datasets, comparing it against established baselines for time-varying optimization: TV-GP-UCB~\citep{bogunovic2016}, W-GP-UCB~\citep{deng2022weighted}, R-GP-UCB and SW-GP-UCB~\citep{DBLP:Zhoujournals/corr/abs-2102-06296}, and SparQ-GP-UCB~\citep{mauduit2025time}. For completeness, we also include the standard GP-UCB algorithm as a static reference. 
	
	In all experiments, the parameters controlling the temporal variations of the function~$\alpha$ (for W-SparQ-GP-UCB) and~$\epsilon$ (for TV-GP-UCB) are tuned on a training dataset. The window size and weighting parameters of R-GP-UCB, SW-GP-UCB, and W-GP-UCB are set according to the recommendations of their respective authors. For W-SparQ-GP-UCB, we fix the window size parameter to~$\tilde{\alpha} = 1/4$.
	
	\subsection{Synthetic data}
	
	We consider the domain~$\Dset = [-50, 50]$ and use the SE kernel
	\begin{align*}
		k : \Dset \times \Dset &\to \mathbb{R}_+, &
		k(x,x') &= 0.5 \exp\!\left(-\frac{\|x-x'\|_2^2}{18}\right),
	\end{align*}
	with variance parameter~$M_k^2 = 0.5$ and lengthscale~$l = 3$. We construct a time-varying function~$f$ belonging to the RKHS associated with~$k$, such that for all~$t \in \mathbb{N}$, $\|f_t\|_{\mathcal{H}_k} \leq 5$. Let~$(c_i)_{i=1}^n \subset \Dset$ be a set of kernel centers and~$(a_i(t))_{i=1}^n$ be time-dependent coefficients. Then,
	\begin{align*}
		f : \Dset \times \mathbb{N} &\to \mathbb{R}, &
		f(x,t) &= \sum_{i=1}^n a_i(t) \, k(x, c_i),
	\end{align*}
	belongs to~$\mathcal{H}_k$ as a finite linear combination of kernel evaluations. Denoting~$K_C = [k(c_i, c_j)]_{i,j}$, the RKHS norm satisfies
	$$
	\|f_t\|_{\mathcal{H}_k}^2 = \boldsymbol{a}(t)^\top K_C \, \boldsymbol{a}(t).
	$$
	We set~$\boldsymbol{a}(t) = \frac{5}{\lambda_{\max}} \frac{\mathbf{u}(t)}{\|\mathbf{u}(t)\|}$, where~$\mathbf{u}_i(t) = \sin(0.3t + i)$ and~$\lambda_{\max}$ is the largest eigenvalue of~$K_C$. This ensures~$\|f_t\|_{\mathcal{H}_k} \leq 5$ and provides smooth temporal variations of the function.
	
	For all seven methods, we plot the mean and standard deviation (across $30$ realizations) of the cumulative regret over~$T = 500$ iterations, along with the theoretical asymptotic upper bound given by Equation~\eqref{Eq:WSparQUpper}.
	
	\begin{figure}
		\centering
		\includegraphics[scale=.5]{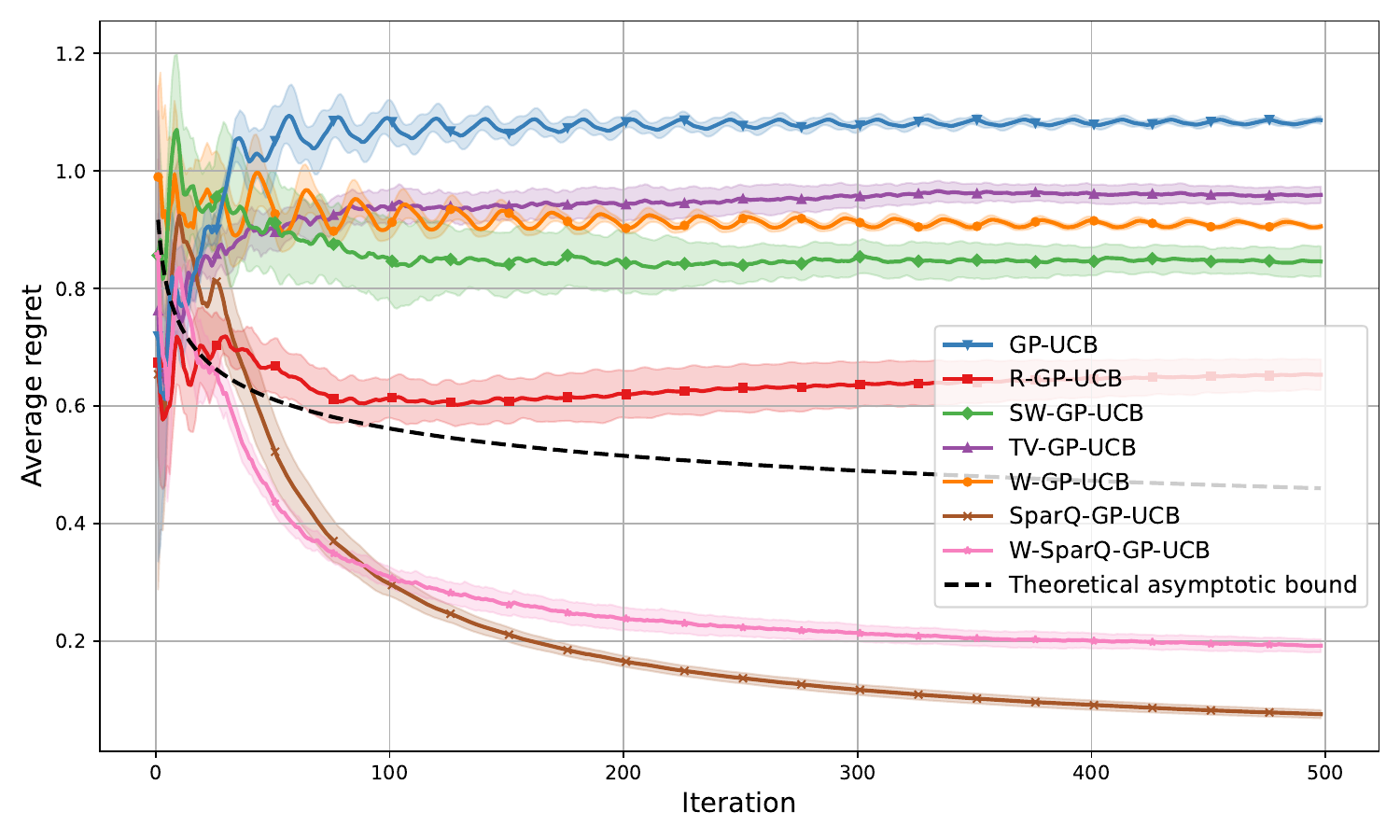}
		\caption{Average regret of GP-UCB variants in the time-varying setting.}
		\label{Fig:RegretStats}
	\end{figure}
	
	As predicted by the theoretical analysis, both SparQ-GP-UCB and W-SparQ-GP-UCB achieve sublinear cumulative regret and outperform existing bandit algorithms. Although SparQ-GP-UCB converges slightly faster, it requires significantly more additional queries ($\mathcal{O}(T \log^d T)$ against~$\mathcal{O}(T^{-\tilde{\alpha}/\alpha} \log^d T)$ for W-SparQ-GP-UCB). This illustrates the expected trade-off between convergence speed and query efficiency established in our theoretical bounds.
	
	\subsection{Real data}
	
\begin{figure}
		\centering
		\includegraphics[scale=.5]{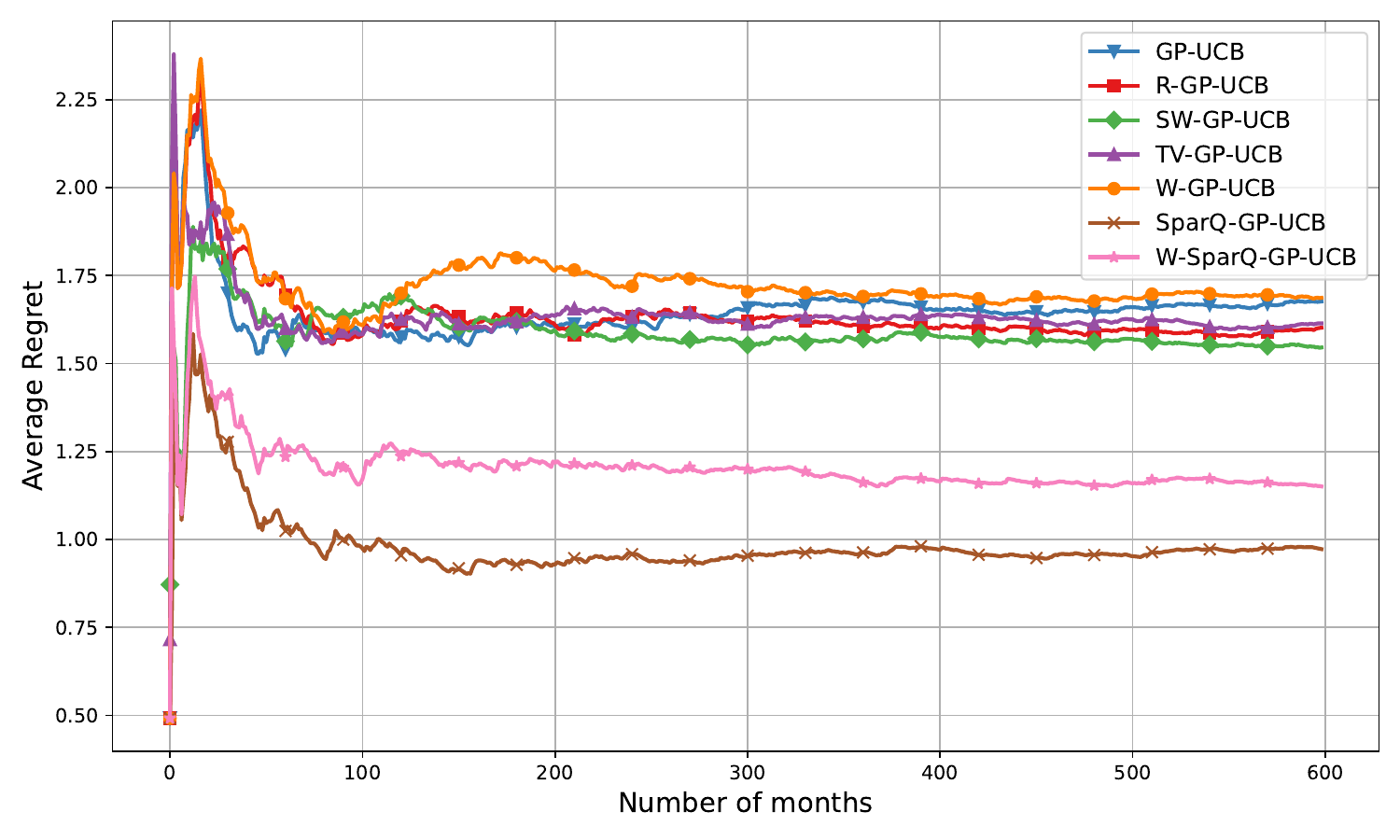}
		\caption{Average regret of GP-UCB variants on Berkeley Earth dataset.}
		\label{Fig:RegretReal}
	\end{figure}

	We evaluated our methods on a real-world dataset, using the \textit{Berkeley Earth High-Resolution (Beta)} dataset~\citep{BerkeleyEarthData}, which provides a gridded time series of global monthly temperature anomalies starting from 1850. For our experiments, we restricted the analysis to monthly temperature anomalies in France from October 1975 to October 2025. The average regret of the seven methods is shown in Figure~\ref{Fig:RegretReal}.
	
\begin{figure}
		\centering
		\includegraphics[scale=.5, trim=1cm 1.5cm 1cm 5mm, clip=on]{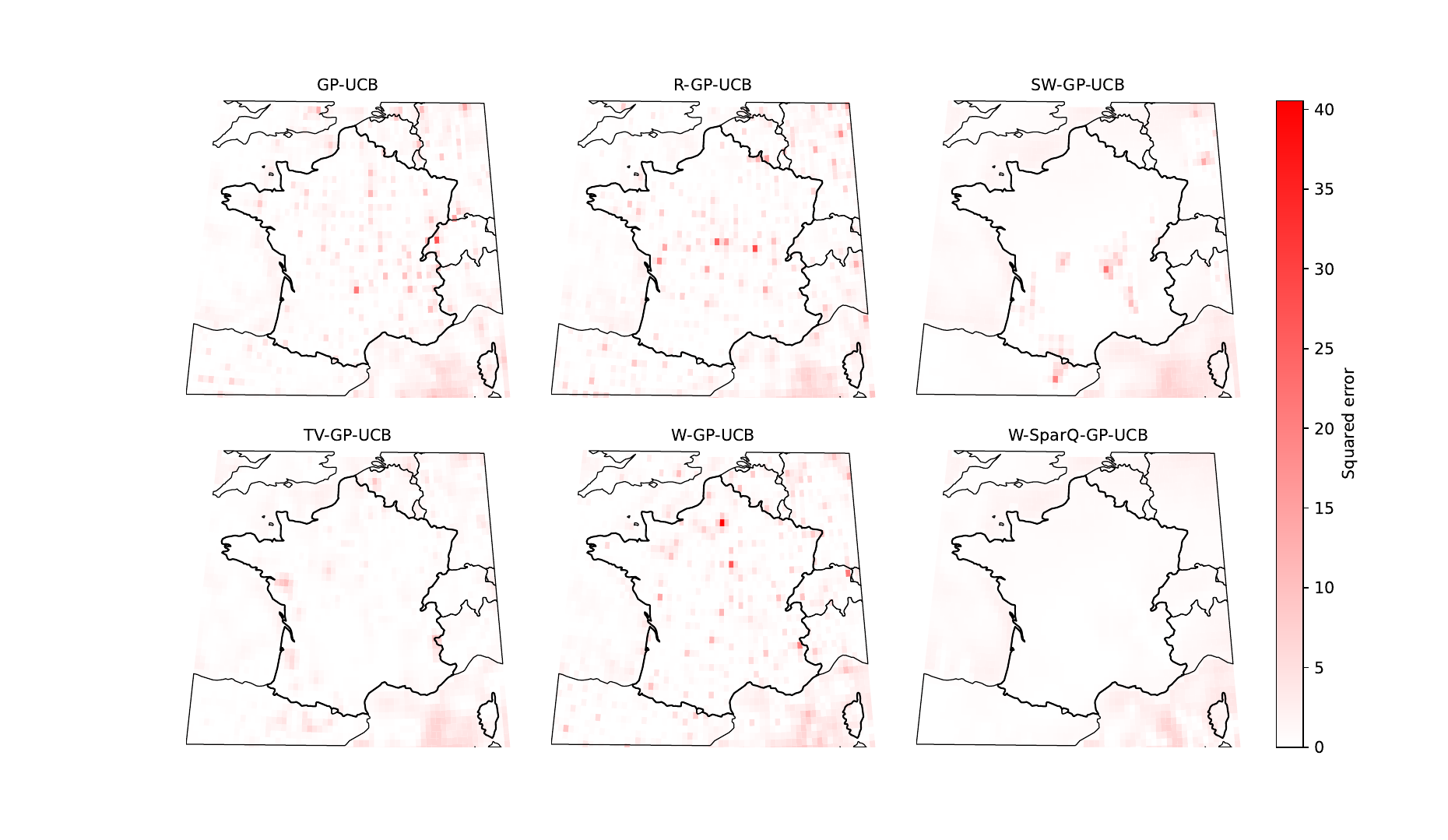}
		\caption{Squared prediction error of GP-UCB variants.}
		\label{Fig:PredError}
	\end{figure}

	On this real-world dataset, although SparQ-based methods outperform the other GP-UCB variants, they all appear to exhibit linear regret. While this may seem at odds with the theoretical results, several factors can explain this discrepancy. First and foremost, the true underlying function governing monthly temperature anomalies may not lie in the RKHS associated with the SE kernel. Even if it did, the hyperparameters, such as the kernel bound~$M_k$, the length-scale~$l$, and~$\alpha$, were estimated using standard procedures (maximization of the log marginal likelihood on a training set), and are likely inaccurate. Indeed, these methods are highly sensitive to hyperparameter tuning.
	In contrast, for the synthetic-data experiments, the time-varying function was constructed using the exact parameters, eliminating the need for hyperparameter training and explaining the matching-theory performance. 
	To further assess performance, we computed the squared prediction error over the entire spatial grid. The results are shown in Figure~\ref{Fig:PredError}.

	These results indicate that the methods can be used not only for optimization but also for prediction tasks. This can be attributed to the fact that UCB methods promote broad exploration and thus yield accurate global predictions. Indeed, the theory guarantees that the true underlying function belong to the confidence interval over the whole domain~$\Dset$, not only around the optimizer. The maps highlight that W-SparQ-GP-UCB outperforms the other bandit baselines in real-world conditions, producing lower and smoother prediction errors.

	\section{Lower bounds on the number of queries in fast variations regimes}\label{sec:LowerBoundQueries}
	In this section, we derive a qualitative lower bound on the number of additional queries required to achieve sublinear regret when~$\alpha \geq 1$. Then, we refine this result when~$\alpha > 1$ by providing a qualitative lower bound. For moderate variations regimes, i.e.~$1/3<\alpha<1$, standard tools used in the proofs of this section fail to provide necessary conditions. Intuitively, when the observation noise increases sufficiently fast over time, too few additional queries prevent reliable identification of the optimum, resulting in linear regret.
	\paragraph{Setting and notations}
	Fix a time horizon~$T \in \mathbb{N}$.
	We consider a Gaussian process model with a SE kernel~$k$ and observations following a time-varying noise model:
	\begin{equation}\label{eq:tv-noise-recap}
		y_t = f_t(x_t) + \epsilon_t, \qquad
		\epsilon_t \sim \mathrm{subG}\big(0, \sigma^2(1 + (T-t)^{\alpha})\big),
	\end{equation}
	where~$\alpha > 0$ controls the rate at which the observation noise increases as we move backward in time from the horizon~$T$. 
	At each step~$t$, the online algorithm may perform~$n_t \ge 0$ additional queries\footnote{This time, not necessarily at previously queried points.} in addition to its main evaluation at~$x_t$. Let
	$$
	N_T=\sum_{t=1}^T n_t,
	$$
	denote the total number of additional queries up to time~$T$.
	We denote by~$\mathbb{E}[R_T]$ the expected cumulative regret of the algorithm over the horizon~$T$.
	We study the problem of optimizing a different function~$f_t : \Dset \to \mathbb{R}$ at each step~$t$, where the sequence~$(f_t)_{t \ge 1}$ satisfies Assumptions~\ref{as.1} and~\ref{as.2}. Consequently, at time~$T$, every past observation~$(x_t, y_t)$ follows the time-varying model~\eqref{eq:tv-noise-recap}.
	The class of algorithms considered here differs from that used in Section~\ref{Sec:BanditLowerBounds}:
	we allow online policies~$\pi$ that, at each time~$t$, select both the main evaluation point~$x_t$ and the~$n_t$ additional queries while enforcing $\sum_{t=1}^T n_t =N_T$. Thus, at time~$T$, a total of~$T + N_T$ observations have been collected.
	We aim at establishing minimax lower bounds on the expected regret~$\mathbb{E}[R_T]$ for this class of problems and algorithms, and to characterize their dependence on~$N_T$.
	If, for a given~$N_T$, the corresponding regret lower bound is (super)linear in~$T$, this value of~$N_T$ constitutes a necessary lower bound on the number of additional queries required to achieve sublinear regret.
	We begin with a qualitative argument that introduces the main intuition and analytical tools used to derive such bounds, before presenting a quantitative result that explicitly relates the minimal required number of additional queries to the noise growth parameter~$\alpha$.
	
	\subsection{Qualitative bound on~$N_T$}\label{Sec:QualBound}
	
	We first present a qualitative lower bound showing that a bounded number of additional queries cannot prevent linear regret in rapidly varying regimes.
	
	\begin{proposition}\label{Prop:QualitativeQueriesLowerBound}
		Let the sequence~$(f_t)_{t\ge1}$ satisfy Assumptions~\ref{as.1}--\ref{as.2}, with~$\alpha \geq 1$, and let the regression kernel be squared-exponential. Fix a horizon~$T$. Consider any online algorithm~$\pi$ that may perform a total of~$N_T$ additional queries until step~$T$. If the sequence~$(N_T)_{T\ge0}$ is bounded (i.e., there exists~$C$ such that~$N_T\le C$ for all~$T$), then the expected cumulative regret satisfies
		$$
		\mathbb{E}[R_T] \geq \Omega(T).
		$$
		In other words, a uniformly bounded total number of additional queries is insufficient to guarantee sublinear regret.
	\end{proposition}

    \begin{proof}
    If the increasing sequence~$(N_T)_{T \geq 0}$ is bounded, then there exists~$T_1$ such that, for all~$T \geq T_1$,~$N_T = N_{T_1}$. In other words, from iteration~$T_1$, we recover the bandit feedback and we can apply Proposition~\ref{Prop:BanditLowerRegret} and~$R_T = \Omega(T)$. 
    \end{proof}
    Hence, it is necessary that the number of additional queries goes to infinity when the number of iteration goes to infinity to achieve sublinear regret. We specify this result in the following section by providing a quantitative bound for the necessary number of additional queries.

	\subsection{Quantitative bound on~$N_T$}
	
	The following theorem gives a necessary scaling for $N_T$ to avoid linear regret.
	
	\begin{theorem}\label{Thm:LowerBoundSideQueries:formal}
		Under the same hypothesis as Proposition~\ref{Prop:QualitativeQueriesLowerBound}, if
		\[
		N_T = \mathcal{O}\!\big(T^{\frac{\alpha}{\alpha+1}}\big),
		\]
		then the expected cumulative regret satisfies
		\[
		\mathbb E[R_T] \ge \Omega(T).
		\]
		In particular, any algorithm with~$N_T = \mathcal{O}(T^{\alpha/(\alpha+1)})$ side queries cannot guarantee sublinear regret.
	\end{theorem}
	
	\begin{proof}
		We work at a frozen horizon~$T$ and establish a lower bound on~$\mathbb{E}[R_T]$ for the class of problems and algorithms depicted above.

		Fix a small amplitude~$\gamma>0$. We consider the adversary class~$\mathcal{F}_M = \{f^1, \ldots, f^M\}$, constructed as in~\cite{scarlett2017lower}, Sections~3 and~4. The~$f^m$s are translations of scaled inverse Fourier transforms of the multi-dimensional bump functions:
	\[
		H(\xi) = \left\{
		\begin{array}{ll}
			\exp \left( - \frac{1}{1- \|\xi\|_2^2} \right) &\mbox{if } \|\xi\|_2 <1 \\
			0 & \mbox{else.}
		\end{array}
		\right.
	\]
		Each~$f^m$ attains a unique maximum of height~$2\gamma$ located at the center of a distinct region~$\mathcal R_m$ of a uniform partition of~$\Dset$. Figure~\ref{Fig:BumpFunctions} should be taken as an illustration for dimension~$d=1$.
		
		\begin{figure}
			\centering
			\includegraphics[scale=.8]{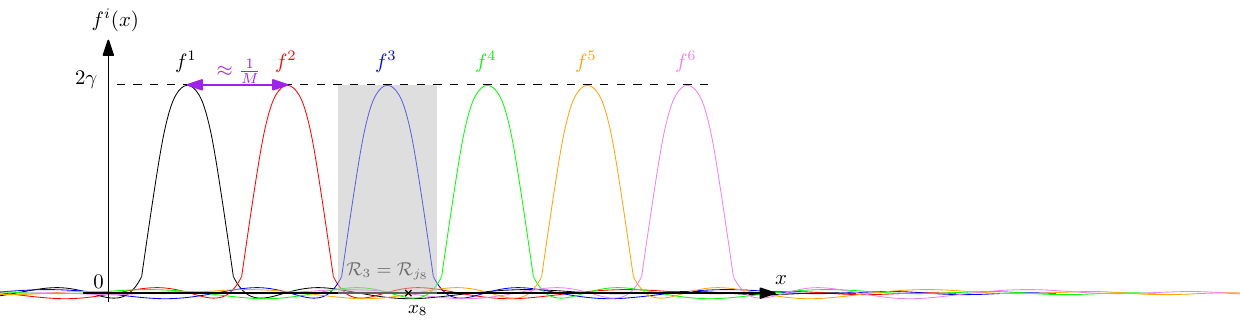}
			\caption{Schematic illustration of six adversary functions for $d=1$.}
			\label{Fig:BumpFunctions}
		\end{figure}

		The construction ensures that any~$\gamma$-optimal point for~$f^i$ cannot be~$\gamma$-optimal for~$f^j$ when~$i\ne j$, and that the RKHS norms~$\|f^m\|_k$ are uniformly bounded by a prescribed~$B$ provided~$M$ and~$\gamma$ are chosen appropriately (see~\cite{scarlett2017lower}, Section~IV.B). They show that there exists a trade-off between the number of functions~$M$ in the adversary set~$\mathcal{F}_M$ and the amplitude~$\gamma$ such that the RKHS norms of each~$f^m \in \mathcal{F}_M$ is bounded by~$B$. The relation between~$M$ and~$\gamma$ is:
		\begin{equation}\label{Eq:MGamma}
			M = \left \lfloor \left(\frac{\sqrt{\log \frac{B (2 \pi l^2)^{d/4}h(0)}{2 \gamma}}}{l \pi \zeta}\right)^d \right \rfloor,
		\end{equation}
		where~$l$ is the lengthscale of the kernel,~$h$ is the inverse Fourier transform of~$H$ and~$\zeta$ is a constant such that, for~$\|x\|_\infty > \zeta$,~$h(x) < \frac{h(0)}{2}$.
		We assume the unknown target~$f_T$ equals a uniformly randomly sampled element of~$\mathcal F_M$. Let~$P_i$ denote the distribution of the observed data (including both algorithm-chosen observations and side queries) when the real underlying function~$f_T$ is~$f_T=f^i$. By Fano's lemma (see Theorem~1 in~\cite{scarlett2019introductory}) applied to the problem of identifying the index~$m$, the probability of error~$P_{\mathrm{err}}$ of any estimator~$\hat{m}$ satisfies:
		\begin{equation}\label{eq:fano-basic}
			P_{\mathrm{err}} \ge 1 - \frac{\frac{1}{M^2}\sum_{i\ne j}\mathrm{KL}(P_i\|P_j) + \log 2}{\log M}.
		\end{equation}
		For the~$T+N_T$ conditionally Gaussian observations obtained at horizon~$T$, denoted as~$(\tilde{X}_T, \tilde{Y}_T)$ with noise variances~$(\tilde{\sigma}^2_t)_{t=1}^{T+N_T}$, we have
		\[
		\mathrm{KL}(P_i\|P_j) = \sum_{t=1}^{T+N_T} \frac{\big(f^i(x_t)-f^j(x_t)\big)^2}{2\,\tilde\sigma_t^2}.
		\]
		Summing over~$i\ne j$ and noticing
		\begin{align*}
			\sum_{i=1}^M \sum_{j=1}^M \left(f^i(x_t) - f^j(x_t) \right)^2 &= 2M \sum_{i=1}^M f^i(x_t)^2 - 2 \left( \sum_{i=1}^M f^i(x_t)\right)^2  \leq 2M \sum_{i=1}^M f^i(x_t)^2,
		\end{align*}
		we obtain
		\[
		\sum_{i\ne j}\mathrm{KL}(P_i\|P_j) \le \sum_{t=1}^{T+N_T} \frac{M}{\tilde\sigma_t^2}\sum_{i=1}^M \max_{x\in\mathcal R_{j_t}} \big(f^i(x)\big)^2,
		\]
		where~$j_t$ is the index of the partition cell containing~$x_t$. By Lemma~5 of~\cite{scarlett2017lower}, the inner sum is~$\mathcal O(\gamma^2)$. Therefore there exists~$C_1>0$ (independent of~$T+N_T$) such that
		\begin{equation}\label{eq:KL-sum-bound}
			\sum_{i\ne j}\mathrm{KL}(P_i\|P_j) \leq C_1\,M\gamma^2 \sum_{t=1}^{T+N_T} \frac{1}{\tilde\sigma_t^2}.
		\end{equation}
		The observations~$(\tilde{X}_T, \tilde{Y}_T)$ can be separated into two subsets of observations~$(\tilde{X}_T, \tilde{Y}_T)_{t=1}^T$ (main evaluations) and~$(\tilde{X}_T, \tilde{Y}_T)_{t=T+1}^{T+N_T}$ (additional queries), if necessary by reordering the sets. Hence
		\begin{equation}\label{Eq:InvVarSum}
			\sum_{t=1}^{T+N_T} \frac{1}{\tilde\sigma_t^2} = \underbrace{\sum_{t=1}^T \frac{1}{\sigma^2(1+(T-t)^\alpha)}}_{\leq C_2 \mbox{ independent of } T \mbox{ as } \alpha >1} + \underbrace{\sum_{t=T+1}^{T+N_T} \frac{1}{\tilde\sigma^2_t}}_{= \sum_{t=1}^T \frac{n_t}{\sigma^2(1+(T-t)^\alpha)}}.
		\end{equation}
    
        Now, in order to capture the cumulative information provided by additional queries up to a reference time~$\tau$, we introduce the \emph{information sum}:
		\begin{equation}\label{eq:info-sum}
			S_\tau = \sum_{t=1}^\tau \frac{n_t}{\sigma^2(1+(\tau-t)^\alpha)},
		\end{equation}
		so that, at time~$\tau$ we have
		\begin{equation}\label{Eq:InvVarSum2}
			\sum_{t=1}^{\tau + N_{\tau}} \frac{1}{\tilde\sigma_t^2} \leq C_2 + S_{\tau}.
		\end{equation}
		This way, if~$S_\tau$ remains uniformly bounded by a constant (independent of~$T$), then the right hand side of~\eqref{eq:KL-sum-bound} at time~$\tau$ is bounded by a constant, so by~\eqref{eq:fano-basic} the error probability~$P_{\mathrm{err}}$ (still at time~$\tau$) is bounded away from zero, independently from~$T$.
		
		Now, the objective is to show that $N_T = \mathcal{O}\!\big(T^{\frac{\alpha}{\alpha+1}}\big)$ implies the existence of ``many'' times~$\tau_k \leq T$ with bounded~$S_{\tau_k}$. Intuitively, such~$\tau_k$ are the endpoints of ``empty'' temporal windows in terms of additional queries. Hence, these windows carry little information, inducing high instantaneous regret.
		Fix an integer window length~$1\le L\le T/2$ and consider the~$T-L+1$ overlapping sliding windows~$\mathcal{S}_j = [j+1, j+L]$ for~$j=0,\ldots,T-L$. Let~$w_j=\sum_{t=j+1}^{j+L} n_t$ be the number of additional queries inside window~$\mathcal{S}_j$. Then, the average number of additional queries per sliding window equals
		\[
		\bar N^{(L)} := \frac{1}{T-L+1}\sum_{j=0}^{T-L} w_j = \frac{L}{T-L+1} \sum_{t=1}^T n_t = \frac{L\,N_T}{T-L+1}.
		\]
		The different elements are illustrated in Figure~\ref{Fig:LowerBoundProof}.
		Since~$L\le T/2$, we have~$\bar N^{(L)}\le 2L N_T/T$. By a pigeonhole-type reasoning, there are at least~$m_T\ge \lfloor(T-L+1)/2\rfloor$ sliding windows containing at most~$2\bar N^{(L)} \le 4L N_T/T$ queries.
		Let~$\{\tau_i\}_{i=1}^{m_T}$ be the right endpoints of those windows. For any such~$\tau=\tau_i$ the information sum~\eqref{eq:info-sum} obeys
		\[
		S_\tau \le \frac{1}{\sigma^2}\left(\underbrace{\frac{N_T}{L^\alpha}}_{\text{outside } \mathcal{S}_i} + \underbrace{\frac{4L N_T}{T}}_{\text{inside } \mathcal{S}_i}\right) = \frac{N_T}{\sigma^2}\,F(L),
		\]
		where~$F(L):=L^{-\alpha} + 4L/T$.

		\begin{figure}[H]
		\centering
		\includegraphics[scale=.6]{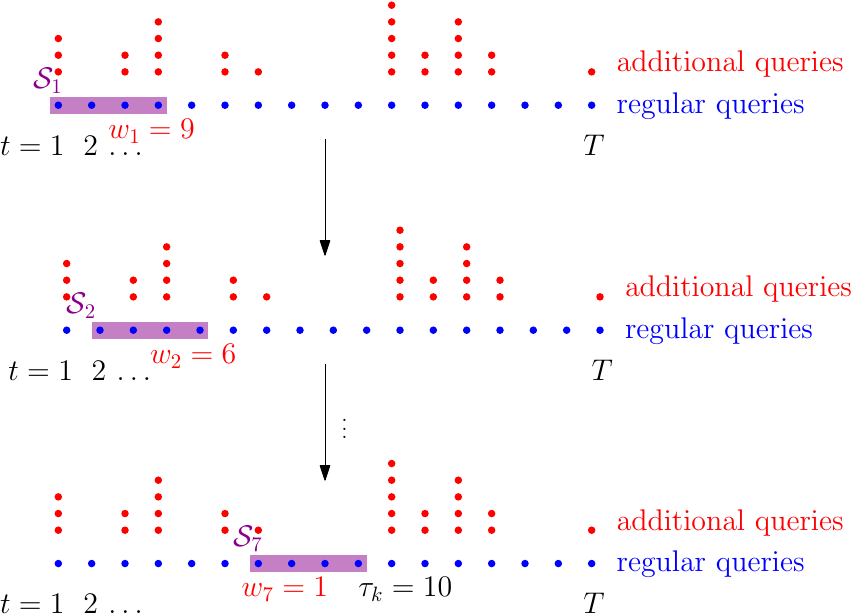}
		\caption{Informativeness of sliding windows. The latter are represented in purple and their size is~$L=4$}
		\label{Fig:LowerBoundProof}
		\end{figure}
		Solving~$F'(L^*)=0$ yields the critical value
		\[
		L^* = \left(\tfrac{\alpha}{8}\right)^{\!1/(\alpha+1)} T^{1/(\alpha+1)},
		\]
		which satisfies~$L^*\le T/2$ for~$T \geq T_0 = 2^{1- \frac{2}{\alpha}} \alpha^\frac{1}{\alpha}$. Substituting~$L^*$ into~$F$ gives the scaling
		\[
		F(L^*) = \Theta\big(T^{-\alpha/(\alpha+1)}\big), \qquad m_T=\Theta(T).
		\]
		Hence, there exists~$C_3$ depending only on \(\alpha,\sigma\) such that
		\[
		\forall i \in \{1, \ldots, m_T\}, \qquad S_{\tau_i} \le C_3\, N_T\, T^{-\alpha/(\alpha+1)}.
		\]
		Now, if~$N_T = \mathcal{O}(T^{\alpha/(\alpha+1)})$, there exists~$C_4>0$ such that~$S_{\tau_i}\le C_4$ uniformly for all~$\tau_i$ and~$T \geq T_0$. Combining this inequality with Equations~\eqref{eq:KL-sum-bound} and~\eqref{Eq:InvVarSum2} we obtain:
		\[
		\sum_{i\ne j}\mathrm{KL}(P_i\|P_j) \leq C_1(C_2+C_4)\,M\gamma^2.
		\]

        Let~$0 < p_0 < 1$ independent of~$T$. As~$M \underset{\gamma \to 0}{\longrightarrow} \infty$ (see Equation~\eqref{Eq:MGamma}),~$P_{\mathrm{err}} \underset{\gamma \to 0}{\longrightarrow} 1$ and there exists an instance~$(\gamma_0, M(\gamma_0))$\footnote{Obtained by solving~$1- \frac{\frac{C}{M(\gamma_0)^2}+\log 2}{\log M(\gamma_0)} = p_0$, with~$0 < p_0 < 1$.} such that~$P_{\mathrm{err}} \geq p_0$ whenever~$\gamma \leq \gamma_0$. Consider the endpoints of ``empty'' windows~$\{\tau_i\}_{i=1}^{m_T}$, and assume that, for any~$i \leq m_T$, there exists~$M_i \leq M$ such that~$f_{\tau_i} = f^{M_i}$ and denote by~$\hat{M}_i$ the estimator of such index.
        Consider the event~$\mathcal{E}_{\tau_i} = \{\hat{M}_i \neq M_i\}$. Then~$P(\mathcal{E}_{\tau_i}) = P_{\mathrm{err}}$, which is bounded independently of~$T$, and
		\begin{align*}
			\mathbb{E}[r_{\tau_i}] &= P(\bar{\mathcal{E}}_{\tau_i}) \mathbb{E}[r_{\tau_i} \vert \bar{\mathcal{E}}_{\tau_i}] +P(\mathcal{E}_{\tau_i}) \mathbb{E}[r_{\tau_i} \vert \mathcal{E}_{\tau_i}] \\
			& \geq P(\mathcal{E}_{\tau_i}) \mathbb{E}[r_{\tau_i} \vert \mathcal{E}_{\tau_i}].
		\end{align*}
		By construction of the family~$\mathcal{F}_M$, on the event~$\mathcal{E}_{\tau_i}$ the algorithm's chosen action at the time when performance is measured cannot be~$\gamma$-optimal for the true function~$f^{M_i}$, so~$\mathbb{E}[r_{\tau_i} \vert \mathcal{E}_{\tau_i}]$ and~$\mathbb{E}[r_{\tau_i}] \geq p_0 \gamma$.
		This strategy can be repeated for every~$\tau_i, \quad i=1,\dots,m_T$, with~$m_T=\Theta(T)$ and finally
		\[
		\mathbb E[R_T]\geq \sum_{i=1}^{m_T}\mathbb E[r_{\tau_i}] \geq \frac{T}{C_3}\,p_0\gamma = \Omega(T),
		\]
		which completes the proof.
	\end{proof}
	
	\begin{remark}[Tightness and interpretation]
		Theorem~\ref{Thm:LowerBoundSideQueries:formal} identifies~${N_T \gg T^{\alpha/(\alpha+1)}}$ as necessary (up to polylogarithmic factors) to avoid linear regret. The upper bounds developed earlier (e.g. Theorem~\ref{Thm:regretW-SparQ}) show that, in favorable regimes, sublinear query schedules exist; together these results suggest the derived scaling is near-optimal, especially when~$\alpha$ is large. Improving constants or removing the gap (polylog factors) would likely require more refined information-theoretic tools (e.g., Le Cam's method to analyze the information brought by additional observations~\citep{lecam1974information}) and is left for future work.
	\end{remark}

	\paragraph{Discussion}
	The lower bound formalizes the intuition that when measurement quality degrades quickly (large~$\alpha$) one must increase the frequency of accurate side observations to keep the information budget large enough to discriminate between close hypotheses. While the qualitative analysis in Section~\ref{Sec:QualBound} sets the intuition, the derived threshold~$N_T \asymp T^{\alpha/(\alpha+1)}$ quantifies this trade-off. Theorem~\ref{Thm:LowerBoundSideQueries:formal} allows to fill the upper left column of Table~\ref{tab:queries_bounds}. For all other configurations, we use the time-invariant bandit lower bound in~\cite{scarlett2017lower}, as not matter how many additional queries we use, time-invariant expected regrets will always be lower than expected regrets in the time-varying configuration.
    Specifically, note that when~$\alpha \leq 1$, then the sum in Equation~\ref{Eq:InvVarSum} is not guaranteed to converge anymore and Fano's lemma becomes trivial (it gives~$P_{\mathrm{err}} \geq 0$).
    
    Closing the remaining polylogarithmic gaps between upper and lower bounds, sharpening constants, and extending the analysis to stochastic function sequences (rather than adversarial families) are natural directions for future work.

	\section{Conclusion}
	
	In this paper, we provide the first characterization of the amount of side information needed for no-regret optimization of time-varying functions with GP bandits. We highlight the two main contributions: W-SparQ-GP-UCB, an efficient algorithm with vanishing queries per iteration, and an analysis of the relationship between the number of additional queries and the regret lower and upper bounds, depending on parameter~$\alpha$.
	Open directions include improving our lower bounds in the intermediate regime~($1/3 \leq \alpha \leq 1$), designing adaptive query strategies, and extending our analysis to other kernels and multi-agent settings.

\section*{Acknowledgements}

The work is partly supported by Hi! Paris and ANR/France 2030 program
(ANR-23-IACL-0005) as well as the ANR project ANR-23-CE48-0011-01.

\bibliographystyle{plainnat}
\bibliography{references}

\end{document}